\documentclass[proceedings]{ccn}
\ccndoi{10.32470/6ff31r0}

\usepackage{hyperref}
\usepackage{url}             
\usepackage{booktabs}        
\usepackage{nicefrac}        
\usepackage{microtype}       
\usepackage{xcolor}          
\usepackage{pifont}          
\usepackage{siunitx}         
\usepackage[normalem]{ulem}  
\usepackage{extarrows}
\usepackage{lipsum}
\usepackage{xspace}

\usepackage{tikz}
\usetikzlibrary{shapes}
\usepackage{dirtytalk}
\usepackage{bbding}         
\usepackage{enumitem}       
\usepackage{empheq}         

\usepackage{tabularray}
\usepackage{tabularx}
\usepackage{makecell}
\usepackage{array}       
\newcolumntype{C}{>{\centering\arraybackslash}X}  

\usepackage{float}
\usepackage{wrapfig}
\usepackage{graphicx}
\graphicspath{{./figs}}

\usepackage{cleveref}
\crefname{figure}{Fig.}{Figs.}
\Crefname{figure}{Figure}{Figures}
\crefname{equation}{eq.}{eqs.}
\Crefname{equation}{Equation}{Equations}
\crefname{table}{Table}{Tables}
\Crefname{table}{Table}{Tables}
\crefname{algorithm}{Alg.}{Algs.}
\Crefname{algorithm}{Algorithm}{Algorithms}

\usepackage{caption}
\usepackage{tcolorbox}
\usepackage[makeroom]{cancel}

\usepackage[utf8]{inputenc}     
\usepackage[T1]{fontenc}        

\usepackage{mathtools,amssymb,amsthm,bm}
\usepackage{cancel}
\usepackage{nicefrac}

\usepackage[textcolor=white,backgroundcolor=red]{todonotes}
\setuptodonotes{inline}

\DeclareRobustCommand{\bx}{\bm{x}}

\DeclareRobustCommand{\bz}{\bm{z}}

\DeclareRobustCommand{\bu}{\bm{u}}
\DeclareRobustCommand{\bv}{\bm{v}}

\DeclareRobustCommand{\bh}{\bm{h}}

\DeclareRobustCommand{\blambda}{\bm{\lambda}}
\DeclareRobustCommand{\btheta}{\bm{\theta}}

\DeclareRobustCommand{\bmu}{\bm{\mu}}
\DeclareRobustCommand{\bsigma}{\bm{\sigma}}

\DeclareRobustCommand{\bphi}{\bm{\phi}}
\DeclareRobustCommand{\bPhi}{\bm{\Phi}}

\DeclareRobustCommand{\cL}{\mathcal{L}}  
\DeclareRobustCommand{\cD}{\mathcal{D}}  
\DeclareRobustCommand{\cF}{\mathcal{F}}  
\DeclareRobustCommand{\cN}{\mathcal{N}}  
\DeclareRobustCommand{\cG}{\mathcal{G}}  
\DeclareRobustCommand{\cP}{\mathcal{P}}  

\DeclareMathOperator{\EE}{\mathbb{E}}
\DeclareMathOperator{\RR}{\mathbb{R}}

\newcommand{\norm}[1]{\left\lVert#1\right\rVert}
\newcommand{\kl}[2]{\cD_{\mathtt{KL}}\!\left({#1}\,\|\,{#2}\right)}
\newcommand{\KL}[2]{\cD_{\mathtt{KL}}\Big{(}{#1}\,\big\|\,{#2}\Big{)}}
\newcommand{\expec}[2]{\EE_{#1} [ {#2} ]}
\newcommand{\expect}[2]{\EE_{#1}\!\Big{[} {#2} \Big{]}}
\newcommand{\EXPECT}[2]{\EE_{#1}\!\Bigg{[} {#2} \Bigg{]}}
\newcommand{\argmin}[1]{\underset{#1}{\operatorname{argmin}}\ }

\def\1{\bm{1}}

\def\vx{{\bm{x}}}

\def\vz{{\bm{z}}}

\DeclareMathAlphabet{\mathsfit}{\encodingdefault}{\sfdefault}{m}{sl}
\SetMathAlphabet{\mathsfit}{bold}{\encodingdefault}{\sfdefault}{bx}{n}

\DeclareMathOperator{\Tr}{Tr}

\DeclareMathOperator{\diag}{diag}
\DeclareMathOperator{\Var}{Var}
\DeclareMathOperator{\Cov}{Cov}

\definecolor{neuro}{HTML}{39A300}
\definecolor{comp}{HTML}{077AF8}
\definecolor{theo}{HTML}{EF4136}
\definecolor{RowColorTop}{HTML}{e0dccb}
\definecolor{RowColorDark}{HTML}{f2efe6}
\definecolor{RowColorLight}{HTML}{fdfbf7}
\definecolor{MSBlue}{rgb}{.204,.353,.541}
\definecolor{MSLightBlue}{rgb}{.31,.506,.741}
\definecolor{BlogBlue}{HTML}{2d6a99}
\definecolor{BlogBG}{HTML}{f9f6f5}
\definecolor{py_0}{HTML}{1F77B4}
\definecolor{py_1}{HTML}{FF7F0E}
\definecolor{py_2}{HTML}{2CA02C}
\definecolor{py_3}{HTML}{D62728}
\definecolor{py_4}{HTML}{9467BD}
\definecolor{py_5}{HTML}{8C564B}
\definecolor{py_6}{HTML}{E377C2}
\definecolor{py_7}{HTML}{7F7F7F}
\definecolor{py_8}{HTML}{BCBD22}
\definecolor{py_9}{HTML}{17BECF}
\definecolor{gr}{HTML}{37C532}
\definecolor{grey}{HTML}{4D4D4D}
\definecolor{gray}{HTML}{4D4D4D}
\definecolor{lightgray}{HTML}{666666}
\definecolor{lightlightgray}{HTML}{999999}
\definecolor{color_enc}{HTML}{ED1C24}
\definecolor{color_dec}{HTML}{29ABE2}
\definecolor{color_decenc}{HTML}{dd1aff}
\def\enc#1{{\color{color_enc}{#1}}\xspace}
\def\dec#1{{\color{color_dec}{#1}}\xspace}

\newcommand{\pbrn}{p_\dec{\text{brain}}}

\newcommand{\pois}{\cP\mathrm{ois}}
\newcommand{\pvae}{$\cP$-VAE\xspace}
\newcommand{\gvae}{$\cG$-VAE\xspace}
\newcommand{\greluvae}{$\cG_{\scriptstyle{\text{relu}}}$-VAE\xspace}

\def\sayit#1{\say{\textit{{#1}}}}

\NewDocumentCommand{\boxit}{ O{} m }{
    \begin{tcolorbox}[
        colback=yellow!20, 
        colframe=lightgray, 
        title={#1} 
    ]
    \centering
    {#2} 
    \end{tcolorbox}
}
\NewDocumentCommand{\circled}{ O{1.8pt} O{0.8pt} m }{
    \tikz[baseline=(char.base)]{
        \node[shape=circle,draw,line width=#2,inner sep=#1] (char) {#3};
    }
}
\newcommand{\underbracegray}[2]{
    {\color{lightlightgray}
        \underbrace{\color{black}#1}_{\color{black}#2}
    }
}
\makeatother
\makeatletter
\newcommand{\namelabel}[1]{%
  \phantomsection
  \renewcommand{\@currentlabel}{#1}%
  \label{#1}%
}
\makeatother
\makeatletter
\RenewDocumentCommand{\boxed}{O{10pt} O{0.8pt} m}{%
  \begingroup
  \setlength{\fboxsep}{#1}%
  \setlength{\fboxrule}{#2}%
  \fbox{$\m@th\displaystyle #3$}%
  \endgroup
}
\makeatother
\makeatletter
\NewDocumentCommand{\boxedcap}{O{1.0pt} O{0.2pt} m}{%
  \begingroup
  \setlength{\fboxsep}{#1}%
  \setlength{\fboxrule}{#2}%
  \fbox{$\m@th\displaystyle #3$}%
  \endgroup
}
\makeatother

\title{Metabolic cost of information processing in\\Poisson variational autoencoders}

\author{
    Hadi Vafaii, \; Jacob L.~Yates \\
    Redwood Center for Theoretical Neuroscience, UC Berkeley \\
    \email{\{vafaii,yates\}@berkeley.edu}
}

\AtBeginShipoutNext{\global\addtolength{\textheight}{6mm}\global\setlength{\footskip}{20pt}}

\begin{document}
\maketitle

\begin{abstract}
Computation in biological systems is fundamentally energy-constrained, yet standard theories of computation treat energy as freely available. Here, we argue that variational free energy minimization under a Poisson assumption offers a principled path toward an energy-aware theory of computation.
Our key observation is that the Kullback-Leibler (KL) divergence term in the Poisson free energy objective becomes proportional to the prior firing rates of model neurons, yielding an emergent metabolic cost term that penalizes high baseline activity. This structure couples an abstract information-theoretic quantity---the \textit{coding rate}---to a concrete biophysical variable---the \textit{firing rate}---which enables a trade-off between coding fidelity and energy expenditure.
Such a coupling arises naturally in the Poisson variational autoencoder (\pvae)---a brain-inspired generative model that encodes inputs as discrete spike counts and recovers a spiking form of \textit{sparse coding} as a special case---but is absent from standard Gaussian VAEs. To demonstrate that this metabolic cost structure is unique to the Poisson formulation, we compare the \pvae against \greluvae, a Gaussian VAE with ReLU rectification applied to latent samples, which controls for the non-negativity constraint. Across a systematic sweep of the KL term weighting coefficient $\beta$ and latent dimensionality, we find that increasing $\beta$ monotonically increases sparsity and reduces average spiking activity in the \pvae. In contrast, \greluvae representations remain unchanged, confirming that the effect is specific to Poisson statistics rather than a byproduct of non-negative representations.
These results establish Poisson variational inference as a promising foundation for a resource-constrained theory of computation.
\end{abstract}

\addtocontents{toc}{\protect\setcounter{tocdepth}{-1}}

\begin{figure*}[t]
\centering
\caption*{
\textbf{The variational free energy equation:} relating \textit{model evidence} (left hand side) to the \textit{variational free energy} objective ($\mathrm{ELBO} = -\cF$), plus the standard \textit{KL objective} (used as the starting point in variational inference). Importantly, the left hand side does not depend on the variational parameters, $\enc{\blambda}$; therefore, minimizing $\cF$ with respect to $\enc{\blambda}$ directly minimizes the original inference KL objective.
In short: $\boxedcap{\,\mathrm{evidence}(\bx; \dec{\btheta}) \;=\; -\cF(\bx; \dec{\btheta}, \enc{\blambda}) \,+\, \mathtt{KL}(\bx; \dec{\btheta}, \enc{\blambda})\,}~.$
}
\vspace{-4mm} 
\begin{equation}\label{eq:main}
\boxed[10pt][0.8pt]{
    \underbracegray{
        \log p(\bx; \dec{\btheta})
    }{\substack{\text{model evidence}\\[1.1mm]\text{(log marginal)}}}
    \;\;\;=\;\;\;\,
    \underbracegray{
        \underbracegray{
            \expect{\bz \sim q_\enc{\blambda}(\bz \vert \bx)}{\log p(\bx \vert \bz; \dec{\btheta})}
        }{\text{Reconstruction term (\textit{accuracy})}}
        \;\,-\;\,
        \underbracegray{
            \KL{q_\enc{\blambda}(\bz \vert \bx)}{p(\bz; \dec{\btheta})}
        }{\text{KL term (\textit{coding rate})}}
    }{
        \mathrm{ELBO}(\bx; \dec{\btheta}, \enc{\blambda})
        \;\;\;=\;\;\, -
        \cF(\bx; \dec{\btheta}, \enc{\blambda})
    }
    \;\;\;+\;\;\,
    \underbracegray{
        \KL{q_\enc{\blambda}(\bz \vert \bx)}{p(\bz \vert \bx; \dec{\btheta})}
    }{\substack{\text{KL term (\textit{original objective})}\\[0.7mm]\text{quantifies inference quality}}}
}
\end{equation}
\vspace{-0.1in} 
\end{figure*}

\section{Introduction}
Modern artificial intelligence (AI) has achieved impressive breakthroughs with no slowdown in sight. However, this achievement comes at a serious cost: mainstream AI models are energetically inefficient, posing a global sustainability threat \citep{hao2025empire}. Training and deploying large-scale models consumes gigawatt-hours, and energy is becoming the unavoidable bottleneck for AI progress \citep{you2025powerfrontier}. This is a fundamental physical constraint we cannot engineer our way around. We urgently need to address this from first principles.

The energy inefficiency of mainstream AI systems originates from a critical design principle: \textit{the decoupling of energy and computation} \citep{landauer1961irreversibility,deacon2011incomplete}. There are no mechanisms internal to architectures like transformers that relate computation to energy expenditure. For a transformer, every token is created equal, as far as voltage goes. This is wasteful by design.

In sharp contrast, energy-efficiency is a core principle of biological computation \citep{sterling2015principles,olshausen1997sparse,quiroga2008sparse}. Brains run on $\sim\!\!\!20$ watts \citep{levy2021communication,Balasubramanian2021BrainPower}, yet perform computations that require megawatt-scale data centers to approximate. This efficiency is likely driven by the \textit{efficient coding hypothesis} \citep{barlow1961possible,barlow1972single,barlow1989unsupervised}, which states that brains adapt to the statistics of their environments \citep{simoncelli2001natural}, minimizing metabolic cost \citep{attwell2001energy, olshausen1996emergence, padamsey2023paying}. A complementary possibility is that the brain's representational form---discrete spiking events---determines the cost structure of the computation itself.

Neuromorphic computing aims to bridge biological and artificial computation \citep{mead2002neuromorphic}, where event-driven architectures like Intel's Loihi \citep{davies2018loihi} already \say{think} in spikes and energy. But we still lack rigorous theoretical foundations to inform future algorithm and hardware co-design. This motivates the need for an energy-aware theory of computation that goes beyond the current frameworks, which are limited to time and space complexity \citep{sipser2012,von1945first}.

\citet{aimone2025neuromorphic} recently approached this from the hardware perspective. He argued that neuromorphic computing has fundamentally different energy scaling compared to the von Neumann architecture. In conventional systems, energy is proportional to total algorithmic work: every operation incurs a fixed cost regardless of what is actually being computed. In neuromorphic systems, energy is instead proportional to the cumulative change of state across the computational graph: if a neuron does not spike and its state does not change, no energy is expended. \citet{aimone2025neuromorphic} further showed that the dominant energy terms all scale with the average firing rate, making sparsity the primary lever for efficiency. However, this analysis addresses only the hardware side. The algorithmic and theoretical foundations remain missing.

Here, we demonstrate how \textit{Poisson variational inference} \citep{vafaii2024pvae,vafaii2025brainlike} naturally leads to the emergence of an energy-aware objective that learns to trade computational accuracy for energy expenditure. We contrast this with standard Gaussian variational inference \citep{kingma2014auto,friston2005theory,friston2009fep,friston2010fep}, revealing that such a metabolic cost term is critically absent from the Gaussian formulation. We provide a theoretical explanation using information geometry \citep{amari2016information}: Poisson and Gaussian distributions have fundamentally different geometries, and only Poisson realizes the kind of energy-computation coupling that \citet{aimone2025neuromorphic} argues for. We then conduct comprehensive experiments that confirm these theoretical predictions.

\subsubsection{Contributions.}
We establish from probabilistic first principles that variational inference under Poisson assumptions naturally produces an emergent metabolic term that \textit{makes silence cheap}, and \textit{couples information rate to firing rate}. This cost structure is strikingly similar to what \citet{aimone2025neuromorphic} arrives at from hardware principles: that energy-efficient neuromorphic computation requires algorithms where energy scales with change-of-state rather than total work, and that sparsity (silence) should be free. The convergence of these two independent lines of reasoning---hardware and information theory---positions Poisson variational inference as a promising foundation for resource-constrained theories of computation that treat energy expenditure as a core consideration.

\section{Background}
\subsubsection{Notation.}
We denote scalars $s \in \RR$ with lowercase letters, vectors $\bv \in \RR^{d}$ with bold lowercase, and matrices $\mathbf{M} \in \RR^{m\times n}$ with bold uppercase.
We use \textcolor{color_enc}{red} / \textcolor{color_dec}{blue} color-coding to indicate the \textcolor{color_enc}{inference} (encoder) / \textcolor{color_dec}{generative} (decoder) components of each model, respectively.

\subsubsection{Perception as inference.}
Brains never have access to ground truth information; therefore, brains must always \textit{infer} the state of the world from partial and often noisy observations. This can be modeled using Bayes' rule:
\begin{equation}\label{eq:bayes_rule}
    \underbracegray{\pbrn(\bz \vert \bx; \dec{\btheta})}{\text{posterior}}    
    \;=\;
    \frac{
        \pbrn(\bx \vert \bz; \dec{\btheta})
        \,
        \pbrn(\bz; \dec{\btheta})
    }{\pbrn(\bx; \dec{\btheta})}
    ~,
\end{equation}
where $\bx \in \RR^M$ are observed data (e.g., images), and $\bz$ are $K$-dimensional latent variables internal to the agent. All distributions $\pbrn(\cdot; \dec{\btheta})$ represent \textcolor{color_dec}{subjective beliefs} internal to the agent, hence the subscript \textcolor{color_dec}{brain}; and $\dec{\btheta}$ are adaptable parameters of these internal models (e.g., synaptic weights in brains, or neural net parameters). To simplify notation, we drop the \textcolor{color_dec}{brain} subscript. But it should be clear from the context (and the blue $\dec{\btheta}$) that we are working with distributions internal to the agent.

\subsubsection{Variational inference and the free energy objective.}
The optimal Bayesian posterior (\cref{eq:bayes_rule}) is often intractable due to the marginal distribution appearing in the denominator, motivating the need for approximations.

Variational inference (VI; \citet{blei2017variational}) offers a powerful framework for approximate inference. In VI, we introduce a distribution $q_\enc{\blambda}(\bz \vert \bx)$, termed the \textit{approximate posterior}, with variational parameters $\enc{\blambda}$ that are optimized to minimize the Kullback-Leibler (KL) divergence:
\begin{equation}\label{eq:vi}
    \enc{\blambda}^*
    \;=\;
    \argmin{\enc{\blambda}}
    \KL{q_\enc{\blambda}(\bz \vert \bx)}{p(\bz \vert \bx; \dec{\btheta})}~.
\end{equation}

However, directly optimizing the KL objective (\cref{eq:vi}) requires access to the intractable posterior (\cref{eq:bayes_rule}). Variational inference circumvents this by instead optimizing an equivalent objective, the \textit{\underline{E}vidence \underline{L}ower \underline{BO}und}, or \textit{variational free energy} ($\mathrm{ELBO} = -\cF$; \cref{eq:main}), which can be constructed and evaluated using only the approximate posterior, the prior, and the likelihood. Maximizing $\mathrm{ELBO}$ (or equivalently, minimizing $\cF$) directly minimizes the original KL objective (\cref{eq:vi}). See appendix for a pedagogical derivation of \cref{eq:main} using simple algebra.



\subsubsection{Distinguishing the two KL terms in the free energy.}
Note how a second KL term appears in the $\mathrm{ELBO}/\cF$ expansion of \cref{eq:main}. This KL component of $\cF$ is different from the original one in \cref{eq:vi} (which was our starting point), and can be interpreted as the information \textit{coding rate}: the additional bits required to encode stimulus-specific information in the posterior beyond what is already captured by the prior \citep{hobson1969new,thomas2006elements, alemi18fixing}.

Hereafter, \say{KL term} refers exclusively to the coding-rate KL within the $\cF$ definition: $\kl{\enc{\mathrm{posterior}}}{\dec{\mathrm{prior}}}$.

\subsubsection{Three distributions, one \texorpdfstring{$\cF$}{F}.}
To fully specify $\cF$ (\cref{eq:main}), we need to choose three distributions: the approximate posterior $q_\enc{\blambda}(\bz \vert \bx)$, the prior $p(\bz; \dec{\btheta})$ and the conditional likelihood $p(\bx \vert \bz; \dec{\btheta})$. In both machine learning and theoretical neuroscience, people often choose Gaussians for all three. However, this choice is typically motivated by convention and computational tractability.

\subsubsection{The latent representational form.}
The choice of prior and the approximate posterior is particularly important, because it determines the \textit{representational form} of the latent space, significantly impacting model behavior and properties. Below, we will demonstrate that choosing Poisson instead of Gaussian leads to intriguing mathematical properties in the free energy objective.

\subsubsection{Poisson variational autoencoder (\texorpdfstring{\pvae}{P-VAE}).}
Brains use discrete spike counts to represent and communicate information. Motivated by this biological reality, \citet{vafaii2024pvae} introduced the \pvae, which replaces the Gaussian latents in a standard Gaussian VAE (\gvae; \cite{kingma2014auto}) with Poisson.

\subsubsection{Variational parameters in VAEs.}
In a \pvae, the variational parameter $\enc{\blambda} \in \RR_{\geqslant 0}^K$ is a $K$-dimensional vector of firing rates, one per model neuron. In a \gvae, to fully specify a (factorized) Gaussian approximate posterior we need both a mean vector $\enc{\bmu} \in \RR^K$, and variance $\enc{\bsigma}^2 \in \RR_{\geqslant 0}^K$. In other words, for a \gvae: $\enc{\blambda} \equiv (\enc{\bmu}, \enc{\bsigma}^2)$.

\subsubsection{Inference with learned encoder neural networks.}
In both \pvae and \gvae, an encoder neural network maps input data to the corresponding distributional parameters required for constructing the approximate posterior $q_\enc{\blambda}(\bz \vert \bx)$. In other words: $\enc{\blambda}(\bx) = \mathrm{enc}(\bx; \enc{\bphi})$, where $\enc{\bphi}$ are parameters of the encoder network. In VAEs, $\dec{\btheta}$ and $\enc{\bphi}$---parameters of the \textcolor{color_dec}{decoder} and \textcolor{color_enc}{encoder} networks---are jointly optimized by minimizing $\cF(\bx; \dec{\btheta}, \enc{\bphi})$ (\cref{eq:main}).

\subsubsection{Residual parameterization in the \texorpdfstring{\pvae}{P-VAE}.}
The \pvae adopts a residual parameterization: instead of producing the full posterior rates $\enc{\blambda}(\bx)$, the encoder network outputs a residual gain modulation, $\enc{\delta \blambda}(\bx)  = \mathrm{enc}(\bx; \enc{\bphi}) \in \RR^K_{\geqslant 0}$, that is element-wise multiplied to learnable prior firing rates $\dec{\blambda}_0 \in \RR^K_{\geqslant 0}$, yielding the final posterior rates:
\begin{equation}\label{eq:residual_poisson}
    \enc{\blambda}(\bx)
    \;=\;
    \dec{\blambda}_0 \odot \enc{\delta \blambda}(\bx)~.
\end{equation}
And the \gvae residual parameterization is defined as:
\begin{equation}\label{eq:residual_gaussian}
    \enc{\bmu}(\bx)
    \,=\,
    \dec{\bmu}_0 + \enc{\delta \bmu}(\bx)~,
    \quad
    \enc{\bsigma}(\bx)
    \,=\,
    \dec{\bsigma}_0 \odot \enc{\delta \bsigma}(\bx)~.
\end{equation}

This residual parameterization is motivated by both computational and biological considerations: maintaining a base firing rate and computing only stimulus-specific residuals is more efficient and is reminiscent of predictive coding \citep{rao1999predictive}, while the multiplicative interaction functionally implements gain control \citep{schwartz2001natural}, similar to the modulation observed in top-down cortical interactions \citep{gilbert2013top}.

In this residual parameterization, the \pvae approximate posterior and prior are given by:
\begin{align}
    q_\enc{\blambda}(\bz \vert \bx)
    \;&=\;
    \pois(\bz; \dec{\blambda}_0 \odot \enc{\delta \blambda}(\bx))~,
    \label{eq:posterior_poisson}
    \\
    p(\bz; \dec{\btheta})
    \;&=\;
    \pois(\bz; \dec{\blambda}_0)~,
    \label{eq:prior_poisson}
\end{align}
where $\pois$ is the Poisson probability density function:
$$
\pois(z; \lambda) \,=\, \frac{\lambda^z e^{-\lambda}}{z!}~.
$$

\section{A metabolic cost term emerges \break in the \texorpdfstring{\pvae}{P-VAE}, but not in \texorpdfstring{\gvae}{G-VAE}}
With this background, we are now ready to introduce the main intriguing property of the \pvae free energy (\cref{eq:main}): the KL term (\textit{coding rate}; \citet{alemi18fixing}) becomes linearly proportional to the prior firing rates $\dec{\blambda}_0$, encouraging the model to maintain a low baseline firing.

In the appendix, we derive the KL term for both \pvae and \gvae. Assuming a single latent dimension ($K = 1$), the final expressions are given by:
\begin{align}
    \text{Poisson}:\quad
    \cD_\mathtt{KL}
    \;&=\;
    \dec{\lambda}_0\, f(\enc{\delta\lambda})~,
    \label{eq:kl_poisson}
    \\
    \text{Gaussian}:\quad
    \cD_\mathtt{KL}
    \;&=\;
    \tfrac{1}{2}\left[
        \tfrac{\enc{\delta\mu}^2}{\dec{\sigma}_0^2}
        + g(\enc{\delta\sigma}^2)
    \right]\!~,
    \label{eq:kl_gaussian}
\end{align}
where $\, f(y) \coloneqq y \log y - y + 1 \,$ and $\, g(y) \coloneqq y - 1 - \log y \,$.

\subsubsection{Quadratic approximation.}
\begin{figure}[ht!]
    \centering
    \includegraphics[width=\columnwidth]{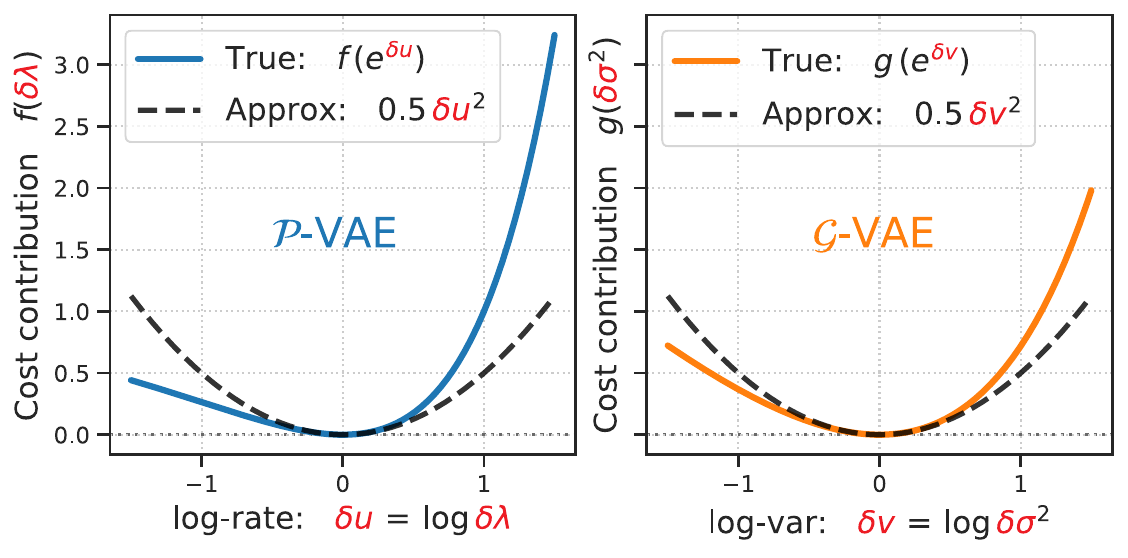}
    \caption{Per-dimension cost contributions of the KL term for the \pvae (left) and \gvae (right), plotted as a function of the log-residual parameters $\enc{\delta u} = \log \enc{\delta\lambda}$ and $\enc{\delta v} = \log \enc{\delta\sigma}^2$, respectively. Solid curves show the exact nonlinearities $f$ (\cref{eq:kl_poisson}) and $g$ (\cref{eq:kl_gaussian}); dashed curves show their quadratic approximations. Both functions share a minimum of zero at the identity residual ($\enc{\delta u} = \enc{\delta v} = 0$, i.e., posterior $=$ prior) and are locally quadratic, but $f$ grows superquadratically for positive deviations, imposing a stronger penalty on firing rate increases than decreases.}
    \label{fig:kl_f_g}
\end{figure}

In the appendix, we compute the leading-order Taylor expansions of the nonlinearities $f(\cdot)$ and $g(\cdot)$, revealing a parallel structure:
\begin{align}
    \text{Poisson}:\quad
    f(\enc{\delta\lambda})
    \;&\approx\;
    \tfrac{1}{2}(\log\enc{\delta\lambda})^2~,
    \label{eq:f_poisson_approx}
    \\
    \text{Gaussian}:\quad
    g(\enc{\delta\sigma}^2)
    \;&\approx\;
    \tfrac{1}{2}(\log\enc{\delta\sigma}^2)^2~.
    \label{eq:g_gaussian_approx}
\end{align}
\Cref{fig:kl_f_g} visualizes $f$ and $g$, and their quadratic approximations. Notably, $f$ grows superquadratically for positive deviations, imposing a stronger penalty on firing rate increases than decreases.

\Cref{eq:kl_poisson,eq:kl_gaussian,eq:f_poisson_approx,eq:g_gaussian_approx} yield the approximate KL terms:
\begin{align}
    \text{Poisson}:\quad
    \cD_\mathtt{KL}
    \;&\approx\;
    \tfrac{1}{2}\,\dec{\lambda}_0\,(\log\enc{\delta\lambda})^2~,
    \label{eq:kl_poisson_approx}
    \\
    \text{Gaussian}:\quad
    \cD_\mathtt{KL}
    \;&\approx\;
    \tfrac{1}{2}
    \frac{\enc{\delta\mu}^2}{\dec{\sigma}_0^2}
    +
    (\log\enc{\delta\sigma})^2~.
    \label{eq:kl_gaussian_approx}
\end{align}

\subsection{KL divergence: Poisson versus Gaussian}
Despite the local similarity (\cref{eq:f_poisson_approx,eq:g_gaussian_approx}), the approximate KL terms (\cref{eq:kl_poisson_approx,eq:kl_gaussian_approx}) reveal two fundamental structural differences.

\subsubsection{The first difference concerns \textit{scaling}.}
For the Poisson, the KL is proportional to the prior rate $\dec{\lambda}_0$; for the Gaussian, the mean-shift cost is scaled by the precision $1/\dec{\sigma}_0^2$, which is independent of the prior mean. This has a direct consequence for what each model considers ``cheap'': the Poisson objective can reduce coding costs by lowering the absolute firing rate ($\dec{\lambda}_0 \to 0$), i.e., \say{silence is cheap}; while the Gaussian can reduce costs by inflating the prior variance ($\dec{\sigma}_0^2 \to \infty$), i.e., \say{ignorance is cheap}. Additionally, the Poisson scaling (\cref{eq:kl_poisson}) is reminiscent of the L1 penalty in sparse coding \citep{olshausen1996emergence}, but with an adaptively modulated penalty strength $f(\enc{\delta\lambda})$ rather than a fixed coefficient.

\subsubsection{The second difference concerns \textit{translation invariance}.}
The Gaussian mean-shift term depends only on the difference $\enc{\delta\mu}^2 = (\enc{\mu} - \dec{\mu}_0)^2$, so the cost is identical whether $\dec{\mu}_0$ is 0 or 1000. In other words, the Gaussian framework has no metabolic floor and is indifferent to the absolute magnitude of activity. In contrast, the Poisson KL breaks this invariance: its linear dependence on $\dec{\lambda}_0$ grounds the computation in a regime where zero activity is the unique lowest-cost state.

\subsubsection{The differences arise from information geometry.}
Both differences are consequences of information geometry \citep{amari2016information}. The KL divergence is locally quadratic in the natural (canonical) parameters of the exponential family. For the Poisson, the natural parameter is $\log\lambda$, and the Fisher information metric in this natural space is $I = \lambda$. This yields a curvature scaled by the rate itself: $\dec{\lambda}_0\,(\log\enc{\delta\lambda})^2$. For the Gaussian mean variable $\mu$, the natural parameter is scaled by variance, but the metric is determined by precision ($I = 1/\sigma^2$), which is, crucially, independent of the mean. This specific coupling between the metric and the parameter magnitude in the Poisson distribution is what gives rise to the emergent metabolic cost (but is absent in the location-scale Gaussian family).

\subsubsection{Summary so far.}
Unlike the Gaussian case, the Poisson KL admits a metabolic cost interpretation, owing to its linear dependence on the prior rate: $\cD_\mathtt{KL} \approx \dec{\lambda}_0(\log\enc{\delta \lambda})^2$:

\namelabel{box:metabolic_cost_interpretation}
\boxit[Poisson KL: a metabolic cost term interpretation]{
To encode information efficiently, maintain low baseline firing rates whenever possible.
}

\subsection{The Poisson KL divergence term couples firing rate to information rate}
The previous section established that the Poisson KL is proportional to the prior firing rate $\dec{\lambda}_0$. We now argue that this mathematical structure carries a deeper significance, by examining what the KL term itself measures.

\subsubsection{Is KL ``just'' a regularizer?}
In much of the machine learning literature, the KL term in the ELBO is treated as a regularizer: a penalty that prevents the approximate posterior from straying too far from the prior, analogous to weight decay or an L2 penalty. Under this view, the KL is just another knob to tune. This framing, while operationally useful, obscures the profound meaning of the KL term. To address this, we ask: \textit{What, precisely, does the KL term quantify in the context of inference?}

\subsubsection{KL as a measure of information gain.}
\citet{hobson1969new} addressed this question from first principles. He was interested in deriving a unique functional that measures the information gained when updating from a prior belief $P$ to a posterior belief $Q$. Starting from a small set of \say{intuitively reasonable properties}---reparameterization invariance, additivity for independent sources, monotonicity, and a composition rule---he proved that the KL divergence is the \textit{only} expression that quantifies information gain while satisfying those intuitive requirements. In sum:

\namelabel{box:hobson}
\boxit[KL as a unique measure of information gain]{
    $\kl{Q}{P}$ is the unique expression for the information contained in a message which alters the probabilities from $P$ to $Q$ \citep{hobson1969new}.
}

\subsubsection{Abstract, information theoretic perspective: KL as information rate.}
This result reframes the KL term from just a regularizer into something far more fundamental: it is the \textit{information rate}---the number of bits per inference step that the stimulus-driven posterior contains beyond what was already present in the prior. Further, KL is not one of many possible measures, or just a definition: it is \sayit{the} measure, uniquely determined by basic desiderata.

\subsubsection{Concrete, biophysical perspective: \texorpdfstring{$\dec{\lambda}_0$}{lambda-0} as metabolic rate.}
Now consider what appears on the right hand side in the Poisson KL expression (\cref{eq:kl_poisson}). The prior rate $\dec{\lambda}_0$ is not an abstract parameter. It represents a concrete biophysical quantity: the baseline number of spikes per unit time. In a potential neuromorphic implementation, it has energy units. As a model of spiking in the brain, it costs ATP (or at least, the synapses responding to those spikes cost ATP \citep{attwell2001energy}). Thus, the prior rate $\dec{\lambda}_0$ is---in this literal sense---a \textit{metabolic rate}.

\subsubsection{Merging the two perspectives:}
The Poisson KL term is a joint cost over information and energy, that couples an abstract \say{Information Rate} (bits / inference step) to a concrete \say{Metabolic Rate} (spikes / inference step):

\namelabel{box:info-couple-metabolic}
\boxit[Poisson KL: Information Rate $\leftrightarrow$ Metabolic Rate]{
\vspace{-4mm}
\begin{equation*}
    \hspace{-3mm}
    \KL{\pois(\bz; \dec{\blambda}_0 \odot \enc{\delta \blambda}(\bx))}{\pois(\bz; \dec{\blambda}_0)}
    \;=\;
    \dec{\blambda}_0^\top f(\enc{\delta\blambda})
\end{equation*}
}

Importantly, the variational free energy objective (\cref{eq:main}) was derived from purely probabilistic considerations---nowhere did we impose an explicit energy constraint. Yet, a metabolic cost term emerged from the mathematics of variational inference under Poisson (but not Gaussian) assumptions.

\subsubsection{Interim conclusion (theory).}
We have theoretically established that the Poisson KL term couples metabolic rate to information rate---a feature with no analog in the Gaussian formulation. We explained this difference through the lens of information geometry, and traced it back to the mean-variance coupling in Poisson.

Beyond mean-variance coupling, another fundamental difference between Poisson and Gaussian distributions is their support: Poisson is nonnegative, whereas Gaussian is unconstrained. Could restricting the Gaussian to positive values induce a similar metabolic cost structure?

\section{The Gaussian-Rectified \texorpdfstring{\greluvae}{Grelu-VAE}}
To control for the positive-support constraint, we apply a ReLU activation to Gaussian samples before passing them to the decoder. The ReLU choice is motivated by several converging lines of evidence: \citet{whittington2023disentanglement} showed that applying ReLU to \gvae, combined with activity penalties, produces disentangled representations that can outperform $\beta$-VAEs \citep{higgins2017beta}; and \citet{Bricken2023SparseFromNoise} showed that combining ReLU with noisy inputs approximates sparse coding.

\subsubsection{The rectification step modifies the decoding pathway.}
\begin{figure}[th!]
    \centering
    \includegraphics[width=\columnwidth]{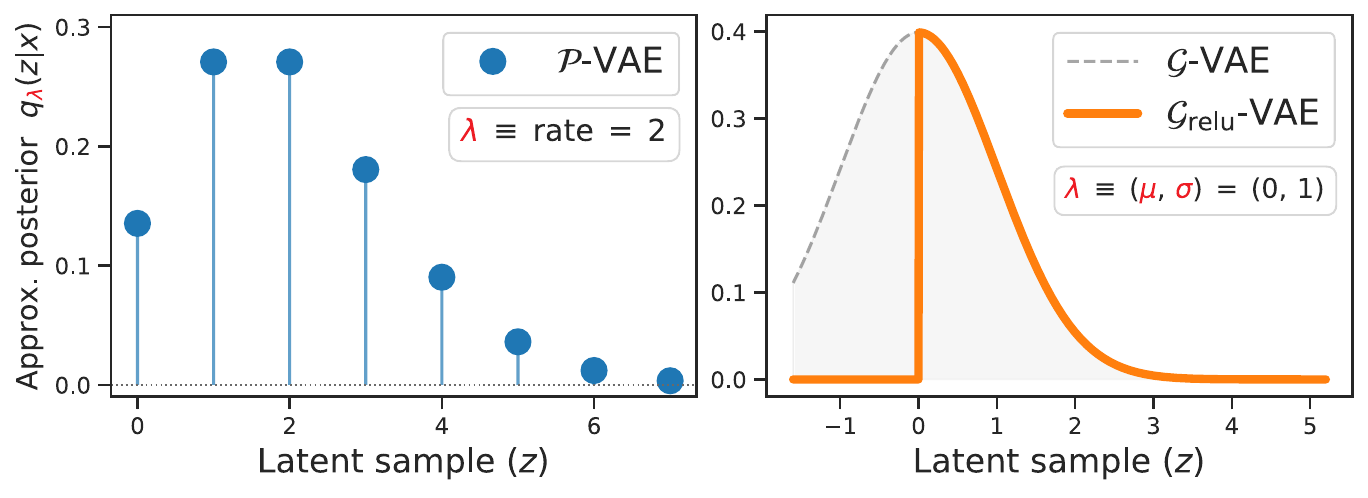}
    \caption{Approximate posterior distributions for a single latent dimension. \textbf{Left:} The \pvae posterior is a Poisson distribution over nonnegative integers (rate $= 2$). \textbf{Right:} The (unnormalized) \greluvae posterior (solid orange) is obtained by applying a ReLU to Gaussian samples (dashed gray, $\mu = 0$, $\sigma = 1$). Both models produce nonnegative latent representations, but through fundamentally different mechanisms: discrete counting statistics versus continuous rectification.}
    \label{fig:dist_pdfs}
    \vspace*{-1mm}
\end{figure}

Given a decoder neural network $\mathrm{dec}(\cdot; \dec{\btheta})$ with parameters $\dec{\btheta}$, the reconstruction is defined as $\hat{\bx} = \mathrm{dec}(\bh; \dec{\btheta})$, where $\bh \equiv \bh(\bz)$ is a function of the latent samples $\bz$:
\begin{align}
    \text{Poisson}:\quad
    \bh \,&=\, \bz~,
    \label{eq:h_poisson}
    \\
    \text{Gaussian}:\quad
    \bh \,&=\, \mathrm{relu}(\bz) \,=\, \max(\bm{0}, \bz)~.
    \label{eq:h_gaussian}
\end{align}
We refer to this resulting rectified Gaussian model as \greluvae. \Cref{fig:dist_pdfs} compares the approximate posterior distributions of \pvae and \greluvae.

\section{Linear decoders, closed-form reconstruction loss, and the variance penalty}
Recall from \cref{eq:main} that the free energy objective decomposes into two terms: $\cF = \cL_\text{recon} + \cD_\mathtt{KL}~.$
So far, we have focused solely on the KL term ($\cD_\mathtt{KL}$), and derived its closed-form expression for both Poisson and Gaussian latent variables (\cref{eq:kl_poisson,eq:kl_gaussian}). We now turn to the reconstruction term.

\subsubsection{The reconstruction term.}
$\cL_\text{recon}$ measures the quality of the latent representation: if the encoding is high fidelity, the data sample $\bx$ will have high expected likelihood under the approximate posterior. Throughout this work, we assume a Gaussian conditional likelihood with $2 \dec{\bsigma}_\text{dec}^2 = \mathbf{1}$, yielding the sum of squared errors reconstruction loss:
\begin{equation}\label{eq:recon_mse}
\begin{aligned}
    \cL_\text{recon}
    \;&=\;
    \expect{\bz \sim q_\enc{\blambda}(\bz \vert \bx)}{-\log p(\bx \vert \bz; \dec{\btheta})}
    \\
    \;&=\;
    \expect{\bz \sim q_\enc{\blambda}(\bz \vert \bx)}{\norm{\bx - \mathrm{dec}(\bz; \dec{\btheta})}_2^2}~,
\end{aligned}
\end{equation}
where $\mathrm{dec}(\bz; \dec{\btheta})$ is the decoder neural network.

\subsubsection{Linear decoders and closed-form reconstruction.}
Consider a linear decoder $\mathrm{dec}(\bz; \dec{\btheta}) = \dec{\bPhi} \bz$, where $\dec{\bPhi} \equiv \dec{\btheta}$ are the decoder parameters. This is a standard choice in theoretical neuroscience \citep{olshausen1996emergence,rao1999predictive}, where $\dec{\bPhi}$ is often called the \textit{dictionary}. In machine learning, \citet{Lucas2019BlameELBO} relied on linear VAEs to study posterior collapse, and \citet{shao2025zerovar} used them to introduce zero-variance gradients for VAEs.

Given a linear decoder, the expectation in \cref{eq:recon_mse} depends only on the first and second moments of $q$, allowing for an analytical derivation (see appendix):
\begin{equation}\label{eq:recon_analytical}
    \cL_\text{recon}
    \;=\;
    \underbracegray{\norm{\bx - \dec{\bPhi}\expec{\enc{q}}{\bz}}_2^2}{\text{mean penalty}}
    \,+\,
    \underbracegray{\diag(\dec{\bPhi}^\top\!\dec{\bPhi})^\top\Var_\enc{q}[\bz]}{\text{variance penalty}}~.
\end{equation}

\subsubsection{The variance penalty.}
The second term in \cref{eq:recon_analytical} penalizes high posterior uncertainty. For Poisson, $\Var_\enc{q}[\bz] = \enc{\blambda}$, linking variance directly to firing rate. This introduces a sparsity pressure distinct from the KL mechanism (\cref{eq:kl_poisson_approx}): dimensions with larger dictionary norms ($[\dec{\bPhi}^\top\!\dec{\bPhi}]_{ii}$) must maintain lower firing rates ($\enc{\lambda}_i$) to minimize free energy. Biologically, this says that neurons with larger receptive fields should fire more sparingly.

For the standard Gaussian, variance is independent of the mean, so no such sparsity interpretation arises. But what about the rectified \greluvae?

\begin{figure}[ht!]
    \centering
    \includegraphics[width=\columnwidth]{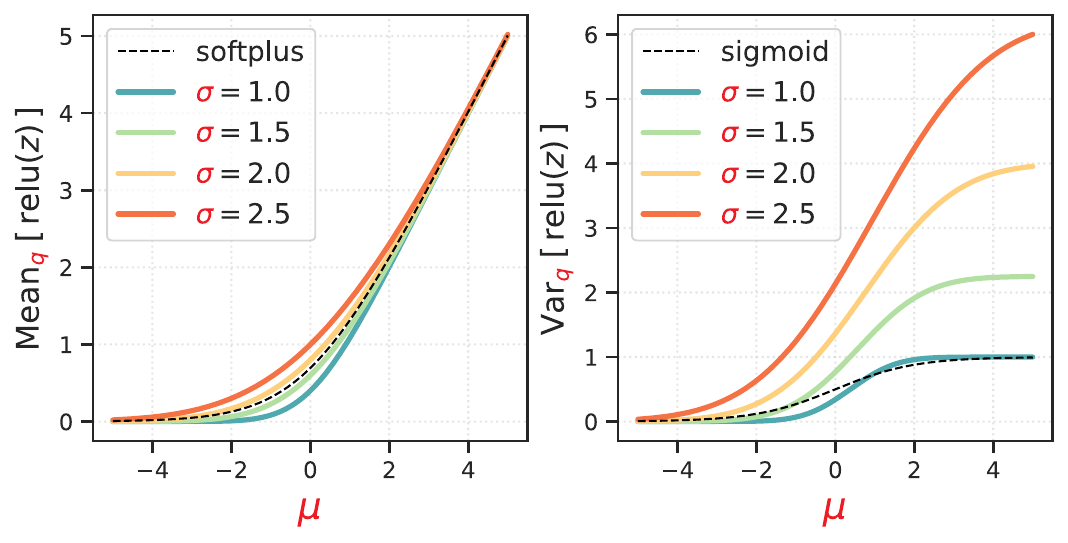}
    \vspace*{-6mm} 
    \caption{Mean (left) and variance (right) of $\mathrm{relu}(z)$ for $z \sim \cN(\enc{\mu}, \enc{\sigma}^2)$, plotted as a function of $\enc{\mu}$ for several values of $\enc{\sigma}$. The mean (\cref{eq:grelu_moments_m}) behaves similar to softplus (dashed), approaching the identity for $\enc{\mu} \gg 0$ and vanishing for $\enc{\mu} \ll 0$. The variance (\cref{eq:grelu_moments_v}) behaves as a sigmoid-like (but steeper) gating function, saturating at $\enc{\sigma}^2$ for $\enc{\mu} \gg 0$ and vanishing for $\enc{\mu} \ll 0$. Crucially, rectification couples mean and variance: pushing $\enc{\mu} \to -\infty$ suppresses both simultaneously, providing a mechanism for the \greluvae to achieve sparse, low-variance representations.}
    \label{fig:rectified_moments}
\end{figure}

For \greluvae, we need the moments of $\bh = \mathrm{relu}(\bz)$. We carry out these calculations in the appendix and present the final results here:
\begin{align}
    \enc{\bm{m}}
    \;&\coloneqq\;
    \expec{q}{\bh}
    \;=\;
    \enc{\bmu} \odot \Phi(\enc{\bm{\zeta}})
    \,+\,
    \enc{\bsigma} \odot \phi(\enc{\bm{\zeta}})~,
    \label{eq:grelu_moments_m}
    \\[1mm]
    \enc{\bv}
    \;&\coloneqq\;
    \Var_{q}[\bh]
    \;=\;
    (\enc{\bmu}^2 + \enc{\bsigma}^2) \odot \Phi(\enc{\bm{\zeta}})
    \;+
    \notag
    \\[-1mm]
    &\qquad\qquad\quad\quad\quad
    (\enc{\bmu} \odot \enc{\bsigma}) \odot \phi(\enc{\bm{\zeta}})
    \,-\,
    \enc{\bm{m}}^2~,
    \label{eq:grelu_moments_v}
\end{align}
where $\enc{\bm{\zeta}} \coloneqq \enc{\bmu} \oslash \enc{\bsigma}$ is the standardized mean, and $\Phi(\cdot)$ and $\phi(\cdot)$ are the standard normal CDF and PDF, respectively. \Cref{fig:rectified_moments} visualizes these moments.

\subsubsection{Rectification promotes sparsity.}
The mean-variance coupling in the \greluvae (\cref{eq:grelu_moments_m,eq:grelu_moments_v}) enables suppressing the variance by pushing the posterior mean $\enc{\mu} \to -\infty$, silencing a latent dimension entirely. This provides the \greluvae with a sparsity mechanism through the reconstruction term.

\subsubsection{Conclusion (theory).}
Both \pvae and \greluvae exhibit variance-driven sparsity pressures through the reconstruction term, but only the Poisson KL directly penalizes metabolic cost. Does this mathematical structure drive energy efficiency in practice?

In the next section, we formulate specific predictions from our theory and test them experimentally.


\begin{table*}[t!]
    \vspace{-.1in} 
    \centering
    \caption{
        \textbf{Poisson VAE (\pvae) and Gaussian-Rectified VAE (\greluvae) architectures}. The linear decoder reconstructs $\hat{\bx} = \dec{\bPhi}\bh$, where $\bh$ is the (possibly rectified) latent representation.
        Within this setup, the closed-form reconstruction loss is given by: $\boxedcap[1.5pt]{ \cL_\text{recon} \;=\; \nicefrac{1}{2}\norm{\bx - \dec{\bPhi} \enc{\bm{m}}}_2^2 \,+\, \diag(\dec{\bPhi}^\top\!\dec{\bPhi})^\top\enc{\bv}}~,$ where $\enc{\bm{m}}$ and $\enc{\bv}$ are the mean and variance of $\bh$ under the approximate posterior $q_\enc{\blambda}(\bz \vert \bx)$. This results in a closed-form free energy objective, $\boxedcap[1.5pt]{\cF \,=\, \cL_\text{recon} \,+\, \beta\cD_\mathtt{KL}}~.$
        For the Poisson model, $\dec{\blambda}_0$ denotes the learnable prior rates and $\enc{\delta\blambda} = \enc{\blambda} \oslash \dec{\blambda}_0$ is the element-wise multiplicative residual, interpreted as gain modulation. For the Gaussian model, $\dec{\bmu}_0, \dec{\bsigma}_0$ denote the prior mean and scale, with additive residual $\enc{\delta\bmu} = \enc{\bmu} - \dec{\bmu}_0$ and multiplicative residual $\enc{\delta\bsigma} = \enc{\bsigma} \oslash \dec{\bsigma}_0$. For the Gaussian moments, we have defined $\enc{\bm{\zeta}} \coloneqq \enc{\bmu} \oslash \enc{\bsigma}$, and $\phi(\cdot)$ and $\Phi(\cdot)$ are the standard normal PDF and CDF, respectively.
    }
    \label{tab:model_comparison}
    \renewcommand{\arraystretch}{1.2}
    \setlength{\tabcolsep}{5pt}
    \begin{tabular}{
        @{}
        >{\centering\arraybackslash}p{1.3cm}
        >{\centering\arraybackslash}p{5.2cm}
        >{\centering\arraybackslash}p{1.8cm}
        >{\centering\arraybackslash}p{4.8cm}
        >{\centering\arraybackslash}p{3.8cm}
        @{}
    }
        \toprule
        \textbf{Model}
        &
        \begin{tabular}[c]{@{}c@{}}
        \textbf{Approx.\ posterior} $q_\enc{\lambda}(z \vert x)$
        \\
        (for a single latent variable)
        \end{tabular}
        &
        \begin{tabular}[c]{@{}c@{}}
            \textbf{Decoder}
            \\
            $\hat{\bx} = \dec{\bPhi}\bh$
        \end{tabular}
        &
        \begin{tabular}[c]{@{}c@{}}
            \textbf{Moments of} $\bh$ \textbf{under} $q_\enc{\blambda}(\bz \vert \bx):$
            \\
            $\enc{\bm{m}} = \expec{q}{\bh}$, \, $\enc{\bv} = \expec{q}{\bh^2} - \enc{\bm{m}}^2$
        \end{tabular}
        &
        \begin{tabular}[c]{@{}c@{}}
        \textbf{KL divergence}, $\cD_\mathtt{KL}$
        \\
        (coding rate)
        \end{tabular}
        \\
        \midrule
        \begin{tabular}[c]{@{}c@{}}
            \pvae
        \end{tabular}
        &
        \begin{tabular}[c]{@{}c@{}}
            $q_\enc{\lambda}(z \vert x) = \tfrac{\enc{\lambda}^{\,z} \, e^{-\enc{\lambda}}}{z!}$
            \\[1.5mm]
            $z \in \mathbb{Z}_{\geqslant 0}$, \; $\enc{\lambda}(x) \in \RR_{\geqslant 0}$
        \end{tabular} &
        $\bh = \bz$ &
        \begin{tabular}[c]{@{}c@{}}
            $\enc{\bm{m}} \,=\, \enc{\blambda}$
            \\[0.0mm]
            $\enc{\bv} \,=\, \enc{\blambda}$
        \end{tabular} &
        \begin{tabular}[c]{@{}c@{}}
            $\cD_\mathtt{KL}
            \,=\,
            \dec{\blambda}_0^\top \, f(\enc{\delta\blambda})$
            \\[1.0mm]
            \footnotesize
            $f(y) \,=\, y \log y - y + 1$
        \end{tabular} \\[2mm]
        \midrule
        \begin{tabular}[c]{@{}c@{}}
            \greluvae
        \end{tabular}
        &
        \begin{tabular}[c]{@{}c@{}}
            $q_{\enc{\mu},\enc{\sigma}^2}(z \vert x)
            =
            \tfrac{1}{\sqrt{2\pi \enc{\sigma}^2}}\exp\Bigl(\!-\tfrac{(z - \enc{\mu})^2}{2\enc{\sigma}^2}\Bigr)$
            \\[2.0mm]
            $z \in \RR$, \; $\enc{\mu}(x) \in \RR$, \; $\enc{\sigma}^2(x) \in \RR_{\geqslant 0}$
        \end{tabular}
        &
        $\bh = \mathrm{relu}(\bz)$
        &
        \begin{tabular}[c]{@{}c@{}}
            $\enc{\bm{m}} \;=\; $
            \footnotesize
            $\enc{\bmu} \odot \Phi(\enc{\bm{\zeta}})
            + \enc{\bsigma} \odot \phi(\enc{\bm{\zeta}})$
            \\[1.5mm]
            $\enc{\bv} \;=\; $
            \footnotesize
            $(\enc{\bmu}^2 + \enc{\bsigma}^2)
            \odot \Phi(\enc{\bm{\zeta}})$
            \\[-1.0mm]
            \footnotesize
            $\quad + \enc{\bmu} \odot \enc{\bsigma}
            \odot \phi(\enc{\bm{\zeta}}) - \enc{\bm{m}}^2$
        \end{tabular}
        &
        \begin{tabular}[c]{@{}c@{}}
            $\cD_\mathtt{KL}
            =
            \tfrac{1}{2} \mathbf{1}^\top
            \!\Bigl[
                \tfrac{\enc{\delta\bmu}^2}{\dec{\bsigma}_0^2}
                + g(\enc{\delta\bsigma}^2)
            \Bigr]$
            \\[2.0mm]
            \footnotesize
            $g(y) \,=\, y - 1 - \log y$
        \end{tabular} \\[2mm]
        \bottomrule
    \end{tabular}
\end{table*}

\section{Empirical validation: the Poisson KL term promotes sparsity}

\subsubsection{Experimental design: the \texorpdfstring{$\beta$}{beta} coefficient.}
To test the metabolic cost interpretation, we introduce a weighting parameter $\beta$ that controls the relative strength of the KL term in the free energy (\cref{eq:main}):
\begin{equation}\label{eq:free_energy_beta}
    \cF \;=\;
    \cL_\text{recon}
    \,+\,
    \beta\,
    \cD_\mathtt{KL},
\end{equation}
which is identical to the $\beta$-VAE framework \citep{higgins2017beta}. By varying $\beta$, we modulate the pressure from the KL term and observe how each model responds.

\paragraph{Hypothesis.}
Since only the Poisson KL acts as a metabolic cost term (\cref{eq:kl_poisson_approx} vs.\ \cref{eq:kl_gaussian_approx}), increasing $\beta$ will systematically increase sparsity only in \pvae.

\subsubsection{The full analytical free energy.}
Assuming a linear decoder, the complete objective is:
\begin{equation}\label{eq:free_energy_full}
\boxed[8pt]{
    \cF
    \;=\;
    \norm{\bx - \dec{\bPhi}\enc{\bm{m}}}_2^2
    \,+\,
    \diag(\dec{\bPhi}^\top\!\dec{\bPhi})^\top\enc{\bv}
    \,+\, \beta\, \cD_\mathtt{KL}
}
\end{equation}
where $\enc{\bm{m}} = \enc{\bv} = \enc{\blambda}$ for \pvae, and $\enc{\bm{m}}$, $\enc{\bv}$ are given by \cref{eq:grelu_moments_m,eq:grelu_moments_v} for \greluvae.

\subsubsection{Comprehensive model comparison.}
In this paper, we work with \pvae and \greluvae architectures. \Cref{tab:model_comparison} provides a comprehensive summary of model properties.

\subsubsection{Methods.}
Following \citet{vafaii2024pvae}, we trained \pvae and \greluvae models on whitened $16 \times 16$ natural image patches from the van Hateren dataset \citep{van1998independent}. For each model family, we trained multiple instances across a grid of latent dimensionalities $K$ and KL weighting coefficients $\beta$ (\cref{eq:free_energy_beta}):
\begin{equation*}
\begin{aligned}
    K \;&\in\;
    \{64, 128, 192, 256, 384, 512, 1024, 2048\},
    \\
    \beta \;&\in\;
    \{0.01, 0.1, 0.5, 1.0, 1.5, 2.0, 4.0, 8.0\}.
\end{aligned}
\end{equation*}

\subsubsection{Architectures.}
Both model families used identical architectures---a linear encoder and decoder with Gaussian likelihood (\cref{eq:recon_mse,eq:recon_analytical})---differing only in their distributional assumptions over the latents.

\subsubsection{Representations.}
After training, we extract representations by passing validation data samples $\bx$ through the encoder to obtain the posterior $q_\enc{\blambda}(\vz \vert \vx)$, from which we sample latent representations $\bz$. We then use $\bz$ to evaluate sparsity and metabolic cost.

\paragraph{Metrics.}
We quantify energy efficiency using two complementary metrics, averaged over the validation set:
\begin{align}
    \text{\footnotesize(Metabolic Cost)}
    \quad
    MC
    \,&\coloneqq\,
    \text{mean activity of } \bz
    \label{eq:metric_mc}
    \\
    \text{\footnotesize(Sparsity)}
    \quad
    PZ
    \,&\coloneqq\,
    \text{proportion of zeros in } \bz
    \label{eq:metric_pz}
\end{align}

\Cref{eq:metric_pz} directly reflects computational energy savings in hardware implementations, where zero-valued activations incur no cost \citep{sze2020efficient, gholami2022survey, Hubara2016BinarizedNN, schuman2022neuromorphic}.

\subsubsection{Result \#1: Only \texorpdfstring{\pvae}{P-VAE} KL works as a metabolic cost.}
\begin{figure}[ht!]
    \centering
    \includegraphics[width=\columnwidth]{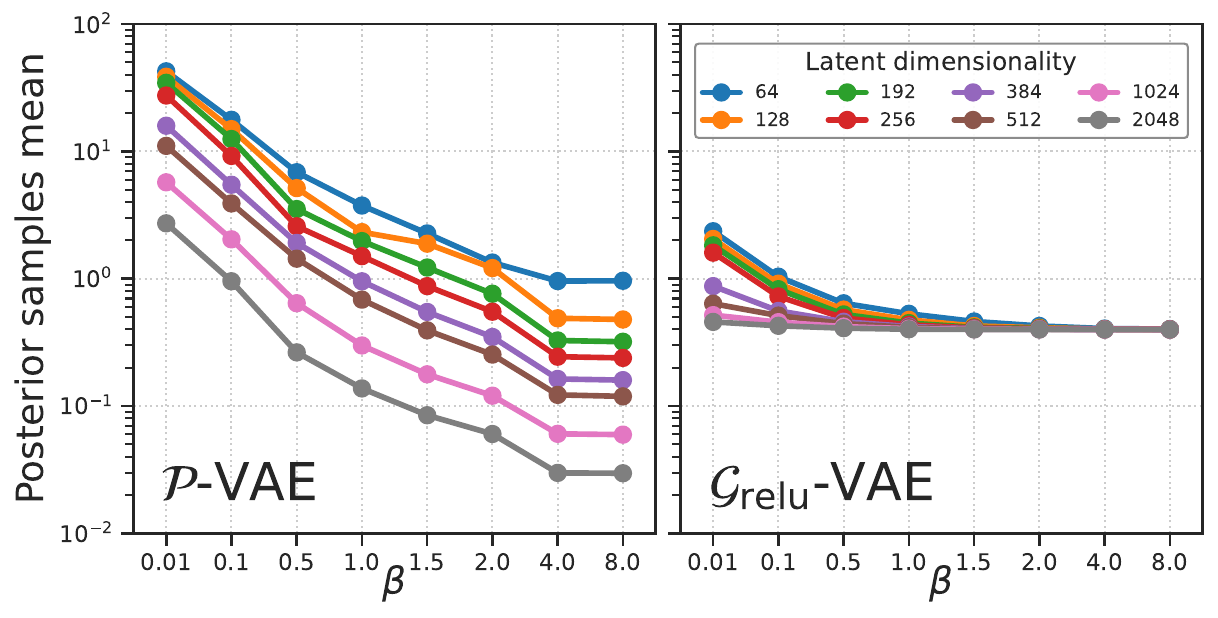}
    \caption{Metabolic cost (\cref{eq:metric_mc}) as a function of KL weighting $\beta$ for different latent dimensionalities. \textbf{Left}: \pvae shows dramatic reduction in metabolic cost with increasing $\beta$, spanning nearly two orders of magnitude. \textbf{Right}: \greluvae shows no systematic change, remaining near baseline across all $\beta$ values. This confirms that the metabolic cost structure is unique to Poisson.}
    \label{fig:posterior_sample_means}
\end{figure}

\Cref{fig:posterior_sample_means} shows the mean neural activity as a function of $\beta$. For \pvae, increasing $\beta$ reduces metabolic cost by nearly two orders of magnitude, confirming that the Poisson KL acts as an effective metabolic cost term whose strength is directly controlled by $\beta$. For \greluvae, mean activity remains flat across all $\beta$ values, confirming the theoretical prediction that the Gaussian KL is decoupled from the magnitude of neural activity.

\subsubsection{Result \#2: Only \texorpdfstring{\pvae}{P-VAE} KL promotes sparsity.}
\begin{figure}[ht!]
    \centering
    \includegraphics[width=\columnwidth]{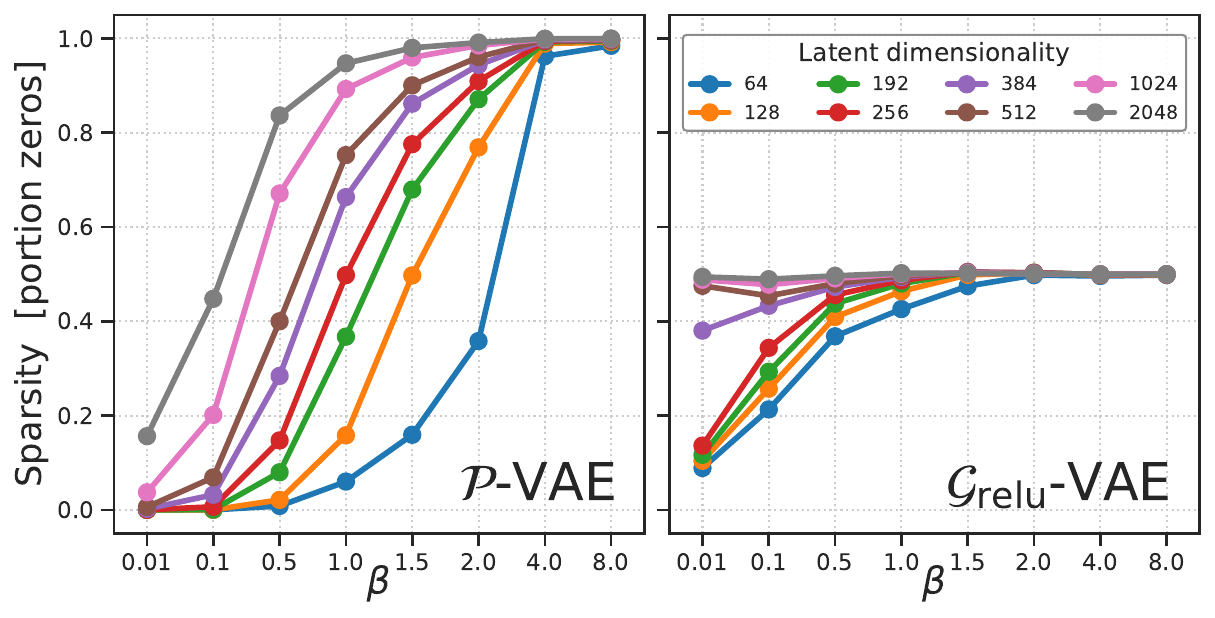}
    \vspace*{-7mm}
    \caption{Proportion of zeros (\cref{eq:metric_pz}) as a function of KL weighting $\beta$ and latent dimensionality $K$. \textbf{Left:} For \pvae, sparsity increases monotonically with $\beta$ across all $K$, with larger $K$ achieving higher sparsity at the same $\beta$. \textbf{Right:} For \greluvae, sparsity plateaus at $50\%$ for all $\beta \geqslant 1.0$, regardless of $K$. See \cref{fig:pie} for a pie chart visualization of the $K=512$ results.}
    \label{fig:portion_zeros}
    \vspace*{-1mm}
\end{figure}

\Cref{fig:portion_zeros} shows sparsity as a function of $\beta$. For \pvae, the proportion of silent neurons increases monotonically with $\beta$ across all latent dimensionalities, reaching near-complete sparsity at high $\beta$.

For \greluvae, sparsity saturates at approximately $50\%$---the expected proportion of negative samples under a symmetric Gaussian---and is insensitive to $\beta$ beyond this level. This $50\%$ ceiling is consistent with the ReLU mechanism: rectification (with pressure from the variance penalty) silences roughly half the neurons by construction, but the Gaussian KL provides no additional pressure to increase sparsity beyond this point.

\subsubsection{Result \#3: \texorpdfstring{\pvae}{P-VAE} and \texorpdfstring{\greluvae}{Grelu-VAE} achieve comparable reconstruction accuracy at matched sparsity, but only Poisson enables a trade-off.}
\begin{figure}[ht!]
    \centering
    \includegraphics[width=\columnwidth]{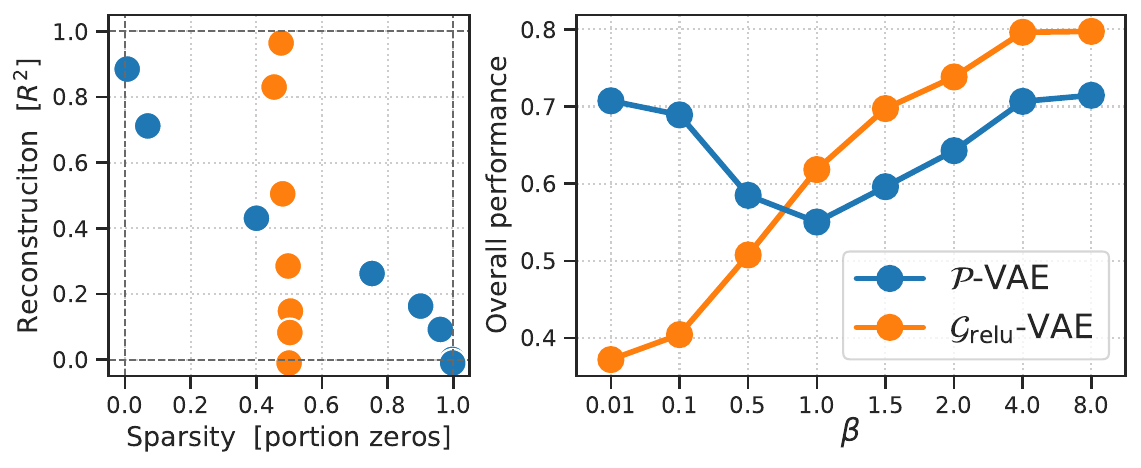}
    \vspace*{-5mm}
    \caption{Reconstruction--sparsity trade-off for $K = 512$. \textbf{Left:} Reconstruction quality ($R^2 \in [-\infty, 1]$) versus sparsity ($PZ \in [0, 1]$; \cref{eq:metric_pz}) across all $\beta$ values. At matched sparsity, both models achieve similar $R^2$, but only \pvae spans the full sparsity range. \textbf{Right:} Overall performance ($\nicefrac{\sqrt{(1 - R^2)^2 + (1 - PZ)^2}}{\sqrt{2}}$; lower is better) as a function of $\beta$. The \pvae achieves its optimum at $\beta = 1$, corresponding to the standard $\mathrm{ELBO}$ (\cref{eq:main}), and exhibits a U-shaped curve reflecting a genuine trade-off between reconstruction and sparsity. In contrast, the \greluvae overall performance degrades monotonically as $\beta$ increases, reflecting that stronger KL weighting hurts reconstruction without yielding additional sparsity gains.}
    \label{fig:rate_dist}
    \vspace*{-3mm}
\end{figure}

\Cref{fig:rate_dist} (left) shows that at matched levels of sparsity, both models achieve comparable reconstruction quality. Therefore, the \pvae does not sacrifice accuracy to become sparse. The key difference is controllability: \pvae admits a smooth trade-off between sparsity and reconstruction, tunable via $\beta$ (\cref{fig:rate_dist}, right). In contrast, \greluvae sparsity is locked near $50\%$ regardless of $\beta$, and increasing $\beta$ worsens reconstruction without affecting sparsity.

\section{Conclusions and Discussion}
We showed that variational inference under Poisson assumptions yields a KL term with a metabolic cost interpretation, enabling a principled trade-off between computational accuracy and energetic cost. This structure is absent in Gaussian formulations---even when rectified---a result we explained using information geometry and verified empirically. Combined with the rate-distortion perspective developed below, these results position Poisson variational inference as a promising foundation for energy-aware theories of computation that could inform algorithm and hardware co-design for efficient AI.

\boxit[Main theoretical takeaways]{
\vspace{-2mm}
\begin{itemize}[itemsep=0pt, leftmargin=-1mm, rightmargin=-3mm]
    \item Gaussian (\textit{ignorance is cheap}): \hfill$\cD_\mathtt{KL} \propto \frac{\enc{\delta\mu}^2}{\dec{\sigma}_0^2} + \dots$
    \vspace{-1.0mm}
    \item Poisson (\textit{silence is cheap}): \hfill$\cD_\mathtt{KL} \propto \dec{\lambda}_0\,(\log\enc{\delta\lambda})^2$
    \item Poisson $\cD_\mathtt{KL}$: \hfill Metabolic Rate $\leftrightarrow$ Information Rate
\end{itemize}
\vspace{-4mm}
}

\subsubsection{The metabolic cost ``emerges'' from the objective, rather than being inserted manually.}
When you write down a free energy, you have to make decisions about which distributions to use. The approximate posterior is particularly important, because it determines the representational form of the latent space. For instance, a Gaussian assumption implies a continuous, unbounded latent space; whereas, a Poisson assumption yields a discrete, integer-valued one, making it more biologically plausible for modeling stochastic spike-count variables. From a more theoretical perspective, Poisson also arises as the maximum-entropy limit of sums of independent Bernoulli events with fixed mean \citep{harremoes2002binomial}, interpretable as spikes driven by many weak synaptic inputs. Together, these arguments make Poisson a principled choice for modeling stochastic spike-count variables.

This motivated \citet{vafaii2024pvae} to introduce the \pvae architecture. They were not looking for a distribution that would yield a metabolic cost. But Poisson variational inference simply ``gifted'' them the metabolic cost term, for free, without ever having to put it there by hand.

\boxit[Metabolic cost from distributional choice]{
A sparsity-inducing metabolic cost penalty simply emerges from choosing Poisson, without having to manually insert it in the loss function.
}

In sharp contrast, for a Gaussian latent space, such a sparsity-inducing term must be bolted on top of the existing free energy objective \citep{whittington2023disentanglement}.

\subsection{Hardware efficiency and algorithm--hardware co-design}
Poisson samples are non-negative integer counts, so zeros are literal absences of events and nonzeros are low-precision event counts. This form is efficient on hardware implementations. On conventional accelerators, energy is dominated not by arithmetic but by data movement \citep{Horowitz2014ComputingEnergyProblem, sze2020efficient}, so the quantities that actually drive cost are the number of active units and the bits spent per value---exactly what a sparse, quantized code minimizes. Such representations cut computation through event skipping, sparse storage, low-precision arithmetic, and reduced memory movement \citep{gholami2022survey, Hubara2016BinarizedNN, sparseAccelerators}. On neuromorphic systems, the alignment is even more promising: energy scales with change-of-state, and a silent neuron costs nothing \citep{aimone2025neuromorphic, davies2018loihi, mead2002neuromorphic, schuman2022neuromorphic}.

Crucially, none of this is bolted on after the fact. There is by now an entire industry devoted to compressing and quantizing dense, continuous models post hoc \citep{gholami2022survey}. In contrast, a Poisson code is sparse and quantized to begin with. The model learns under the very constraints that efficient hardware imposes, rather than being retrofitted to them once training is done.

The point is that energy here is not a fixed constraint the model must satisfy, but a quantity it optimizes adaptively. Because the metabolic cost is the KL term of the very free energy being minimized (\cref{eq:free_energy_beta}), the same gradients that shape the dictionary and reduce reconstruction error also push firing rates down. The model is never told to use less energy: it works out how much to spend, and where, while solving the task, since the spiking activity already counts against the objective. The task and its energy budget are then solved together, at the same time, by a single optimization procedure.

\boxit[Algorithm–hardware co-design implication]{
Poisson variational inference turns energy from an after-the-fact deployment constraint into a variable inside the learning objective: the model learns \textit{what to compute} and \textit{how to budget metabolism} at the same time.
}

\subsection{Connection to rate-distortion theory}
We developed our theory within variational inference \citep{blei2017variational}, but an equivalent formulation arises from \textit{rate-distortion theory} (RDT; \citet{thomas2006elements, alemi18fixing}). In RDT, we minimize a task loss $\cL_\text{task}$ subject to a finite \textit{information rate budget} $\mathcal{B}$. Defining a Lagrange multiplier $\beta$, the objective is:
\begin{equation}\label{eq:RDT}
    \operatorname*{minimize}: \;\;\;\; \cL \,=\, \cL_\text{task} + \beta\, \cD_\mathtt{KL}
    \quad (\text{s.t.} \;\; \cD_\mathtt{KL} \leqslant \mathcal{B})~.
\end{equation}

\Cref{eq:RDT} is mathematically identical to the $\beta$-weighted free energy ($\cF = \cL_\text{recon} + \beta\cD_\mathtt{KL}$; \cref{eq:free_energy_beta}), but with a general task loss in place of reconstruction. Our main contribution is a metabolic cost interpretation of $\cD_\mathtt{KL}$ ($\mathcal{B} \leftrightarrow$ \textit{energy budget}), exclusively under Poisson latent variables. This opens two directions for future work.

\subsubsection{Beyond passive perception.}
Everything we perceive, feel, decide, and do, are ultimately caused by the spiking of neurons in the brain. And it is not unreasonable to model those neurons as Poisson, whatever job they perform. Early in a model's visual pathway, neurons might spike to reconstruct their inputs. In a higher area, they might spike to solve a task, encoded by an optimality variable \citep{levine2018reinforcement}. Downstream in a motor area, neurons might spike to drive movement. What changes across these cases is not the latent distribution, but the objective the spikes are working to optimize.

This is where the rate-distortion view really pays off. Since \cref{eq:RDT} holds for any task loss $\cL_\text{task}$, pairing that loss with the Poisson KL leaves the metabolic-information coupling intact. In this sense, our key result---a metabolic cost interpretation of $\cD_\mathtt{KL}$ ($\mathcal{B} \leftrightarrow$ \textit{energy budget})---is not confined to autoencoding visual input: it reaches any setting whose objective can be written as a well-defined loss, which includes behavior, cognition, and beyond.

\subsubsection{Beyond Poisson.}
But we need not stop at Poisson. Cortical neurons often exhibit super-Poisson variability, well captured by negative binomial distributions \citep{goris2014partitioning}. Crucially, the negative binomial KL divergence is also proportional to the prior firing rate \citep{zhang2025negativebinomial}, modulated by an additional shape parameter. This suggests that the metabolic cost result is not special to Poisson, but a property of count-based latents more generally. Pinning down which information geometries give rise to a metabolic-cost-like KL term is a compelling direction for future work.

\subsection{Energy as the third axis of computation}
Theoretical computer science has historically measured complexity along two axes: \textit{time} and \textit{space} \citep{sipser2012}. Energy expenditure has remained a post-hoc deployment concern. Our results show that the Poisson KL term couples information rate to metabolic rate, so that energy is not an external, fixed constraint, but an intrinsic, dynamic variable of the optimization process.

\subsubsection{An invitation.}
We hope our results will inspire future work toward a more complete, resource-aware theory of computation with energy as its third foundational pillar.

\section*{Acknowledgments}
We would like to thank the Kriegeskorte lab and Frederic Theunissen for inspiring discussions.

\printbibliography

\newpage
\appendix
\onecolumn
\setlength{\parindent}{0pt}
\setlength{\parskip}{6pt} 

\begin{center}
{\Large \textbf{Appendix:\\[2mm]Metabolic cost of information processing in\\Poisson variational autoencoders}}
\end{center}

This appendix provides detailed derivations and supplementary material supporting the main text.

We begin with a pedagogical derivation of the ELBO objective from first principles, requiring only two applications of the multiply-by-one trick, and discuss two complementary decompositions: the rate-distortion view common in machine learning, and the energy-entropy view from statistical physics. We then review the standard Gaussian VAE and survey extensions to non-Gaussian latent distributions.

Next, we derive the closed-form KL divergence for both Poisson and Gaussian distributions, and justify the quadratic approximations used in the main text via second-order Taylor expansion in the natural parameters.

We proceed to derive the analytical reconstruction loss under linear decoders for a general (non-factorized) posterior, and specialize it to the mean-field case.

We then compute the exact first and second moments of the rectified Gaussian distribution, which are required for the closed-form \greluvae objective.

Finally, we provide extended methodological details---including dataset preprocessing, model architectures, training configuration, and compute resources---along with supplementary figures.

\addtocontents{toc}{\protect\setcounter{tocdepth}{2}}
\tableofcontents

\section{Variational inference, the ELBO objective derivation, and VAEs}
Variational inference is a popular approximate inference method that transforms the posterior inference problem into an optimization task \citep{jordan1999introduction, tzikas2008variational, blei2017variational}. Specifically, one introduces a variational density, $q_\enc{\blambda}(\bz \vert \bx)$, parameterized by $\enc{\blambda}$, which approximates the true posterior. The goal is to minimize the Kullback-Leibler (KL) divergence between $q_\enc{\blambda}(\bz \vert \bx)$ and the true posterior $p(\bz \vert \bx; \dec{\btheta})$.

This optimization leads to the standard \textit{\underline{E}vidence \underline{L}ower \underline{BO}und} (ELBO) objective \citep{jordan1999introduction, blei2017variational}, which can be derived starting from the model evidence as follows:
\begin{equation}
\begin{aligned}
    \log p(\bx; \dec{\btheta})
    &\;=\;
    \log p(\bx; \dec{\btheta})
    \underbracegray{\int q_\enc{\blambda}(\bz|\bx) d\bz}{=\,1}
    \\[3mm]
    &\;=\;
    \int q_\enc{\blambda}(\bz|\bx) \log p(\bx; \dec{\btheta}) d\bz
    \\[3mm]
    &\;=\;
    \EXPECT{\bz \sim q_\enc{\blambda}(\bz|\bx)}{\log p(\bx; \dec{\btheta})}
    \\[3mm]
    &\;=\;
    \EXPECT{\bz \sim q_\enc{\blambda}(\bz|\bx)}{\log p(\bx; \dec{\btheta}) + \underbracegray{\log p(\bz \vert \bx; \dec{\btheta}) - \log p(\bz \vert \bx; \dec{\btheta})}{=\,0}}
    \\[3mm]
    &\;=\;
    \EXPECT{\bz \sim q_\enc{\blambda}(\bz|\bx)}{\log \frac{p(\bx; \dec{\btheta}) p(\bz \vert \bx; \dec{\btheta})}{p(\bz \vert \bx; \dec{\btheta})}}
    \\[3mm]
    &\;=\;
    \EXPECT{\bz \sim q_\enc{\blambda}(\bz|\bx)}{\log \frac{p(\bx, \bz; \dec{\btheta})}{p(\bz|\bx; \dec{\btheta})}}
    \\[3mm]
    &\;=\;
    \EXPECT{\bz \sim q_\enc{\blambda}(\bz|\bx)}{\log \frac{p(\bx, \bz; \dec{\btheta}) q_\enc{\blambda}(\bz|\bx)}{p(\bz|\bx; \dec{\btheta}) q_\enc{\blambda}(\bz|\bx)}}
    \\[3mm]
    &\;=\;
    \EXPECT{\bz \sim q_\enc{\blambda}(\bz|\bx)}{\log \frac{p(\bx, \bz; \dec{\btheta})}{q_\enc{\blambda}(\bz|\bx)} + \log \frac{q_\enc{\blambda}(\bz|\bx)}{p(\bz|\bx; \dec{\btheta})}}
    \\[3mm]
    &\;=\;
    \EXPECT{\bz \sim q_\enc{\blambda}(\bz|\bx)}{\log \frac{p(\bx, \bz; \dec{\btheta})}{q_\enc{\blambda}(\bz|\bx)}}
    \,+\,
    \EXPECT{\bz \sim q_\enc{\blambda}(\bz|\bx)}{\log \frac{q_\enc{\blambda}(\bz|\bx)}{p(\bz|\bx; \dec{\btheta})}}
    \\[3mm]
    &\;=\;
    \underbracegray{
        \EXPECT{\bz \sim q_\enc{\blambda}(\bz \vert \bx)}{\log \frac{p(\bx, \bz; \dec{\btheta})}{q_\enc{\blambda}(\bz \vert \bx)}}
    }{
        \mathrm{ELBO}(\bx; \dec{\btheta}, \enc{\blambda})
    }
    \,+\,
    \KL{q_\enc{\blambda}(\bz \vert \bx)}{p(\bz \vert \bx; \dec{\btheta})}.
\end{aligned}
\end{equation}

This concludes our derivation of \cref{eq:main}. All we did was multiply by one twice (once by inserting $\int\! q_\enc{\blambda}(\bz|\bx)d\bz$ and once by introducing the ratio $\frac{q_\enc{\blambda}(\bz|\bx)}{q_\enc{\blambda}(\bz|\bx)}$) followed by a few algebraic rearrangements.

Importantly, \cref{eq:main} holds for any $q_\enc{\blambda}(\bz \vert \bx)$ that defines a proper probability density. Compared to the typical derivation using Jensen's inequality \citep{jordan1999introduction}, this derivation yields a stronger result \citep{kingma2019introduction}: An exact decomposition of model evidence into ELBO plus KL divergence between the approximate and true posterior.

A direct consequence of this equality is that taking gradients of \cref{eq:main} with respect to $\enc{\blambda}$ shows that maximizing the ELBO minimizes the KL divergence, since the model evidence on the left-hand side is independent of $\enc{\blambda}$. Thus, ELBO maximization improves posterior inference quality, corresponding to more accurate perception under the perception-as-inference framework \citep{helmholtz1867handbuch,alhazen}.

\subsection{Two ways to carve up the ELBO}
\label{sec:appendix:carve-up-ELBO}
There are two common ways of expressing and interpreting the ELBO. The first, more popular in the machine learning community, is the VAE loss decomposition \citep{kingma2019introduction}:
\begin{equation}\label{eq:ELBO-carving-vae}
\begin{aligned}
    \mathrm{ELBO}(\bx; \dec{\btheta}, \enc{\blambda})
    \;&=\;
    \EXPECT{\bz \sim q_\enc{\blambda}(\bz \vert \bx)}{\log \frac{p(\bx \vert \bz; \dec{\btheta})p(\bz; \dec{\btheta})}{q_\enc{\blambda}(\bz \vert \bx)}}
    \\
    \;&=\;
    \expect{\bz \sim q_\enc{\blambda}(\bz \vert \bx)}{\log p(\bx \vert \bz; \dec{\btheta})}
    \,+\,
    \expect{\bz \sim q_\enc{\blambda}(\bz \vert \bx)}{\log \frac{p(\bz; \dec{\btheta})}{q_\enc{\blambda}(\bz \vert \bx)}}
    \\
    \;&=\;
    \underbracegray{
        \color{black}\expect{\bz \sim q_\enc{\blambda}(\bz \vert \bx)}{\log p(\bx \vert \bz; \dec{\btheta})}
    }{\text{Reconstruction term (\textit{distortion})}}
    \,-\,
    \underbracegray{
        \color{black}\KL{q_\enc{\blambda}(\bz \vert \bx)}{p(\bz; \dec{\btheta})}
    }{\text{KL term (\textit{coding rate})}}
    \,.
\end{aligned}
\end{equation}

This view emphasizes reconstruction fidelity and latent space regularization. The reconstruction term can be interpreted as a \textit{distortion measure}, quantifying how well the latent code $\bz$ can explain the input $\bx$ through the generative model. The KL term, by contrast, acts as an \textit{information rate}, measuring how much input-dependent information is encoded in the posterior $q_\enc{\blambda}(\bz \vert \bx)$ beyond what is already present in the prior $p(\bz; \dec{\btheta})$. In other words, the KL quantifies the \textit{coding cost} of representing $\bx$ via $\bz$, and reflects the capacity of the latent space to capture novel, stimulus-specific structure. This interpretation is closely related to classical \textit{rate-distortion theory} \citep{thomas2006elements, tishby2000infobn}, and has been formalized in the context of VAEs by \citet{alemi18fixing}.

The second view, more aligned with theoretical neuroscience and physics, splits (negative) ELBO as:
\begin{equation}\label{eq:ELBO-carving-free-nergy}
    -\mathrm{ELBO}(\bx; \dec{\btheta}, \enc{\blambda})
    \;\equiv\;
    \cF(\bx; \dec{\btheta}, \enc{\blambda})
    \;=\;
    \underbracegray{
        \color{black}\expect{\bz \sim q_\enc{\blambda}(\bz \vert \bx)}{-\log p(\bx, \bz; \dec{\btheta})}
    }{\text{Energy}}
    \,-\,
    \underbracegray{
        \color{black}\mathcal{H}\big[q_\enc{\blambda}(\bz \vert \bx)\big]
    }{\text{Entropy}}
    \,,
\end{equation}
where $\mathcal{H}[q] = -\int q \log q$ is the Shannon entropy.

This carving is analogous to the concept of \textit{Helmholtz free energy} from statistical physics \citep{landau1980statistical, jaynes2003probability, Hinton1993HelmholtzFreeEnergy}, where minimizing free energy involves reducing energy while preserving entropy (i.e., maintaining uncertainty). Below, we will use this decomposition to show that the predictive coding objective of \citet{rao1999predictive} can be directly derived from \cref{eq:ELBO-carving-free-nergy} under specific distributional assumptions.

\subsection{Gaussian Variational Autoencoder (\texorpdfstring{\gvae}{G-VAE})}
\label{sec:appendix:gvae}
The standard Gaussian VAE (\gvae) represents a foundational model in the VAE family, where all three distributions are factorized Gaussians \citep{kingma2019introduction}. To simplify the model, the prior is typically fixed as a standard normal distribution with zero mean and unit variance: $p(\bz; \dec{\btheta}) = \cN(\bm{0}, \bm{1})$. The key innovation of VAEs lies in how they parameterize the approximate posterior and likelihood distributions using neural networks.

\subsection{Extensions and modifications of the \texorpdfstring{\gvae}{G-VAE}}
Since the advent of VAEs \citep{kingma2014auto, rezende2014stochastic}, numerous proposals have extended or modified the standard Gaussian framework. These efforts can be broadly categorized into three main directions: (i) developing more expressive or learnable priors, (ii) replacing the likelihood function with non-Gaussian alternatives, and (iii) altering the latent distribution.

In terms of priors $p(\bz; \dec{\btheta})$, researchers have introduced hierarchical variants \citep{sonderby2016ladder, vahdat2020nvae, child2021vdvae,vafaii2023cnvae}, structured priors such as VampPrior \citep{Tomczak2018VampPrior}, and nonparametric approaches like the stick-breaking process \citep{nalisnick2017stickbreaking}.

For the likelihood function $p_\dec{\btheta}(\bx|\bz)$, alternatives to the standard Gaussian have been proposed to better accommodate binary, count, or highly structured data. Examples include the Bernoulli distribution for binary data \citep{kingma2014auto, LoaizaGanem2019ContinuousBernoulli}, Poisson \citep{pandarinath2018lfads} and negative binomial \citep{Zhao2020NegBinomVAE} for count data, and mixtures of discretized Logistic distributions for natural images \citep{vahdat2020nvae, salimans2017pixelcnn}.

For inference, many have enhanced the expressiveness of the variational posterior $q_\enc{\blambda}(\bz|\bx)$ by applying normalizing flows \citep{Rezende2015NF} or inverse autoregressive flows \citep{Kingma2016IAF}, thereby relaxing the mean-field assumption. Others have replaced the Gaussian latents altogether, exploring alternative distributions such as categorical \citep{jang2017categorical, maddison2017concrete}, Bernoulli \citep{rolfe2017dvae, vahdat2018dvaepp}, Laplace \citep{Park19LaplaceVAE}, Dirichlet \citep{srivastava2017autoencoding}, hyperbolic normal \citep{Mathieu2019PoincareVAE}, von Mises-Fisher \citep{Davidson2022vonMisesFisherVAE}, Student's t \citep{kim2024t3}, and negative binomial \citep{zhang2025negativebinomial}. Finally, the most relevant to us is the Poisson VAE (\pvae; \citet{vafaii2024pvae} which has been been extended with iterative inference \citep{vafaii2025brainlike}, and applied to model psychophysical reaction times in decision making \citep{johnson2025pvaert}.

\section{KL divergence: closed-form derivations for Poisson and Gaussian}
\label{sec:kl_derivations}

In this section, we derive the closed-form KL divergence terms for the Poisson and Gaussian distributions used in the main text (\cref{eq:kl_poisson,eq:kl_gaussian}), and justify the quadratic approximations (\cref{eq:kl_poisson_approx,eq:kl_gaussian_approx}) via Taylor expansion.

\subsection{Poisson KL derivation}

For a single latent dimension, let the approximate posterior be $q_\enc{\lambda}(z \vert \bx) = \pois(z; \enc{\lambda})$ and the prior be $p(z; \dec{\btheta}) = \pois(z; \dec{\lambda}_0)$, where $\enc{\lambda}, \dec{\lambda}_0 \in \RR_{>0}$. The KL divergence is:
\begin{equation}\label{eq:poisson-kl-derivation}
\begin{aligned}
    \KL{\pois(z; \enc{\lambda})}{\pois(z; \dec{\lambda}_0)}
    \;&=\;
    \EXPECT{z \sim q_\enc{\lambda}}{\log\frac{\enc{\lambda}^{z}e^{-\enc{\lambda}} / z!}{\dec{\lambda}_0^{z}e^{-\dec{\lambda}_0} / z!}}
    \\[3mm]
    \;&=\;
    \EXPECT{z \sim q_\enc{\lambda}}{z \log \frac{\enc{\lambda}}{\dec{\lambda}_0} - (\enc{\lambda} - \dec{\lambda}_0)}
    \\[3mm]
    \;&=\;
    \enc{\lambda}\log\frac{\enc{\lambda}}{\dec{\lambda}_0} - \enc{\lambda} + \dec{\lambda}_0
    \\[3mm]
    \;&=\;
    \dec{\lambda}_0 \left( \frac{\enc{\lambda}}{\dec{\lambda}_0} \log\frac{\enc{\lambda}}{\dec{\lambda}_0} - \frac{\enc{\lambda}}{\dec{\lambda}_0} + 1 \right)
    \\[3mm]
    \;&=\;
    \dec{\lambda}_0 \, f(\enc{\delta \lambda})~,
\end{aligned}
\end{equation}
where we substituted the expected count $\expec{q}{z} = \enc{\lambda}$, and identified the multiplicative residual $\enc{\delta \lambda} \coloneqq \enc{\lambda}/\dec{\lambda}_0$ and the function $f(y) \coloneqq y \log y - y + 1$.

For a $K$-dimensional latent space, we have:
\begin{equation}
\boxed{
    \KL{\pois(\bz; \dec{\blambda}_0 \odot \enc{\delta \blambda}(\bx))}{\pois(\bz; \dec{\blambda}_0)}
    \;\;=\;\;
    \sum_{i=1}^K
    \dec{\lambda}_{0i}
    f(\enc{\delta\lambda}_i)
    \;\;=\;\;
    \dec{\blambda}_0^\top f(\enc{\delta\blambda})
}
\end{equation}
where $f(\cdot)$ is applied element-wise.

\subsection{Gaussian KL derivation}

For a single latent dimension, let the approximate posterior be $q_{\enc{\mu}, \enc{\sigma}^2}(z \vert \bx) = \cN(z; \enc{\mu}, \enc{\sigma}^2)$ and the prior be $p(z; \dec{\btheta}) = \cN(z; \dec{\mu}_0, \dec{\sigma}_0^2)$. The KL divergence is:
\begin{equation}\label{eq:gaussian-kl-derivation}
\begin{aligned}
    \KL{\cN(z; \enc{\mu}, \enc{\sigma}^2)}{\cN(z; \dec{\mu}_0, \dec{\sigma}_0^2)}
    \;&=\;
    \EXPECT{z \sim q}{\log\frac{\frac{1}{\sqrt{2\pi\enc{\sigma}^2}}\exp\left(-\frac{(z-\enc{\mu})^2}{2\enc{\sigma}^2}\right)}{\frac{1}{\sqrt{2\pi\dec{\sigma}_0^2}}\exp\left(-\frac{(z-\dec{\mu}_0)^2}{2\dec{\sigma}_0^2}\right)}}
    \\[3mm]
    \;&=\;
    \EXPECT{z \sim q}{
        \log\frac{\dec{\sigma}_0}{\enc{\sigma}}
        - \frac{(z-\enc{\mu})^2}{2\enc{\sigma}^2}
        + \frac{(z-\dec{\mu}_0)^2}{2\dec{\sigma}_0^2}
    }
    \\[3mm]
    \;&=\;
    -\frac{1}{2}\log\frac{\enc{\sigma}^2}{\dec{\sigma}_0^2}
    \,-\, \frac{1}{2} \underbracegray{\EXPECT{z \sim q}{\frac{(z-\enc{\mu})^2}{\enc{\sigma}^2}}}{1}
    \,+\, \frac{1}{2\dec{\sigma}_0^2} \EXPECT{z \sim q}{(z - \dec{\mu}_0)^2}
    \\[3mm]
    \;&=\;
    -\frac{1}{2}\log\frac{\enc{\sigma}^2}{\dec{\sigma}_0^2}
    \,-\, \frac{1}{2}
    \,+\, \frac{1}{2\dec{\sigma}_0^2} \Bigg[ \underbracegray{\Var_q[z]}{\enc{\sigma}^2} + (\underbracegray{\expec{q}{z}}{\enc{\mu}} - \dec{\mu}_0)^2 \Bigg]
    \\[3mm]
    \;&=\;
    \frac{1}{2} \left[
        \frac{(\enc{\mu} - \dec{\mu}_0)^2}{\dec{\sigma}_0^2}
        + \left( \frac{\enc{\sigma}^2}{\dec{\sigma}_0^2} - 1 - \log \frac{\enc{\sigma}^2}{\dec{\sigma}_0^2} \right)
    \right]
    \\[3mm]
    \;&=\;
    \frac{1}{2} \left[
        \frac{\enc{\delta\mu}^2}{\dec{\sigma}_0^2}
        + g(\enc{\delta\sigma}^2)
    \right]~,
\end{aligned}
\end{equation}
where we used the standard result $\expec{q}{(z - \dec{\mu}_0)^2} = \Var_q[z] + (\expec{q}{z} - \dec{\mu}_0)^2$, substituted the residuals $\enc{\delta\mu} = \enc{\mu} - \dec{\mu}_0$ and $\enc{\delta\sigma} = \enc{\sigma}/\dec{\sigma}_0$, and identified the function $g(y) \coloneqq y - 1 - \log y$.

For a $K$-dimensional latent space with a factorized (diagonal) posterior, we have:
\begin{equation}
\boxed{
    \KL{\cN(z; \enc{\bmu}, \enc{\bsigma}^2)}{\cN(z; \dec{\bmu}_0, \dec{\bsigma}_0^2)}
    \;\;=\;\;
    \frac{1}{2} \sum_{i=1}^K
    \left[
        \frac{\enc{\delta\mu}_i^2}{\dec{\sigma}_{0i}^2}
        + g(\enc{\delta\sigma}_i^2)
    \right]
    \;\;=\;\;
    \frac{1}{2} \, \mathbf{1}^\top
    \left[
        \frac{\enc{\delta\bmu}^2}{\dec{\bsigma}_{0}^2}
        + g(\enc{\delta\bsigma}^2)
    \right]
}
\end{equation}
where $g(\cdot)$ is applied element-wise, and the division in the precision-weighted term is understood to be element-wise.

\subsection{KL Taylor expansions: Poisson and Gaussian}

Here we derive the quadratic approximations for the nonlinear cost functions $f(\cdot)$ and $g(\cdot)$ used in the main text.

\subsubsection{Poisson: expansion in log-rates}
Let $u \coloneqq \log \enc{\delta\lambda}$ denote the log-residual of the firing rate. Then $\enc{\delta\lambda} = e^u$. Substituting this into the definition of $f(\enc{\delta\lambda})$:
\begin{equation}
    f(e^u) \;=\; e^u \log(e^u) - e^u + 1 \;=\; u e^u - e^u + 1~.
\end{equation}
We perform a Taylor expansion around $u=0$ (corresponding to $\enc{\delta\lambda}=1$, i.e., posterior equals prior):
\begin{align}
    F(u) \;&\coloneqq\; u e^u - e^u + 1 \\
    F'(u) \;&=\; (e^u + u e^u) - e^u \;=\; u e^u \\
    F''(u) \;&=\; e^u + u e^u~.
\end{align}
Evaluating at $u=0$, we find $F(0) = 0$, $F'(0) = 0$, and $F''(0) = 1$. Thus, the second-order approximation is:
\begin{equation}
\boxed{
    f(\enc{\delta\lambda}) \;\approx\; \frac{1}{2} u^2 \;=\; \frac{1}{2} (\log \enc{\delta\lambda})^2
}
\end{equation}

\subsubsection{Gaussian: expansion in log-variance}
Let $v \coloneqq \log (\enc{\delta\sigma}^2)$ denote the log-residual of the \textit{variance}. Then the variance ratio is $\enc{\delta\sigma}^2 = e^v$. Substituting this into the definition of $g(\enc{\delta\sigma}^2)$:
\begin{equation}
    g(e^v) \;=\; e^v - 1 - \log(e^v) \;=\; e^v - 1 - v~.
\end{equation}
We perform a Taylor expansion around $v=0$ (corresponding to matched variances, $\enc{\delta\sigma}^2=1$):
\begin{align}
    G(v) \;&\coloneqq\; e^v - 1 - v \\
    G'(v) \;&=\; e^v - 1 \\
    G''(v) \;&=\; e^v~.
\end{align}
Evaluating at $v=0$, we find $G(0) = 0$, $G'(0) = 0$, and $G''(0) = 1$. Thus, the second-order approximation is:
\begin{equation}
\boxed{
    g(\enc{\delta\sigma}^2) \;\approx\; \frac{1}{2} v^2 \;=\; \frac{1}{2} (\log \enc{\delta\sigma}^2)^2
}
\end{equation}

\section{Derivation of the closed-form reconstruction loss for linear decoders}
\label{sec:linear_decoder_deriv}

In this section, we derive the analytical expression for the reconstruction loss term $\cL_\text{recon}$ (\cref{eq:recon_mse}) under the assumption of a linear decoder.

Let the decoder be defined as $\hat{\bx} = \dec{\bPhi}\bz$, where $\dec{\bPhi} \in \RR^{M \times K}$ is the dictionary matrix. We assume the likelihood is Gaussian with identity covariance (up to a scalar), leading to the mean squared error (MSE) loss:
\begin{equation}
    \cL_\text{recon}
    \;=\;
    \expect{\bz \sim q}{\norm{\bx - \dec{\bPhi}\bz}_2^2}~.
\end{equation}

We expand the squared Euclidean norm:
\begin{equation}
\begin{aligned}
    \norm{\bx - \dec{\bPhi}\bz}_2^2
    \;&=\;
    (\bx - \dec{\bPhi}\bz)^\top (\bx - \dec{\bPhi}\bz)
    \\
    \;&=\;
    \bx^\top\bx \,-\, \bx^\top\dec{\bPhi}\bz \,-\, (\dec{\bPhi}\bz)^\top\bx \,+\, (\dec{\bPhi}\bz)^\top(\dec{\bPhi}\bz)
    \\
    \;&=\;
    \bx^\top\bx \,-\, 2\bx^\top\dec{\bPhi}\bz \,+\, \bz^\top\dec{\bPhi}^\top\dec{\bPhi}\bz~.
\end{aligned}
\end{equation}

Now we take the expectation with respect to the approximate posterior $q_\enc{\blambda}(\bz \vert \bx)$. By linearity of expectation:
\begin{equation}
    \expect{\bz \sim q}{\norm{\bx - \dec{\bPhi}\bz}_2^2}
    \;=\;
    \bx^\top\bx
    \,-\, 2\bx^\top\dec{\bPhi}\expec{\bz \sim q}{\bz}
    \,+\, \expect{\bz \sim q}{\bz^\top\dec{\bPhi}^\top\dec{\bPhi}\bz}~.
\end{equation}

Let $\enc{\bm{m}} \coloneqq \expec{q}{\bz}$ be the posterior mean. The second term simply becomes $-2\bx^\top\dec{\bPhi}\enc{\bm{m}}$. For the third term (quadratic form), we use the trace trick ($\bx^\top \mathbf{A} \bx = \Tr(\mathbf{A}\bx\bx^\top)$):
\begin{equation}
\begin{aligned}
    \expect{\bz \sim q}{\bz^\top\dec{\bPhi}^\top\dec{\bPhi}\bz}
    \;&=\;
    \expect{\bz \sim q}{\Tr(\dec{\bPhi}^\top\dec{\bPhi}\bz\bz^\top)}
    \\
    \;&=\;
    \Tr\left( \dec{\bPhi}^\top\dec{\bPhi} \expect{\bz \sim q}{\bz\bz^\top} \right)~.
\end{aligned}
\end{equation}

Recall the definition of the covariance matrix: $\Cov_q[\bz] = \expec{}{\bz\bz^\top} - \enc{\bm{m}}\enc{\bm{m}}^\top$. Thus, $\expec{}{\bz\bz^\top} = \Cov_q[\bz] + \enc{\bm{m}}\enc{\bm{m}}^\top$. Substituting this back:
\begin{equation}
\begin{aligned}
    \text{Quadratic term}
    \;&=\;
    \Tr\left( \dec{\bPhi}^\top\dec{\bPhi} (\Cov_q[\bz] + \enc{\bm{m}}\enc{\bm{m}}^\top) \right)
    \\
    \;&=\;
    \Tr\left( \dec{\bPhi}^\top\dec{\bPhi} \Cov_q[\bz] \right)
    \,+\,
    \Tr\left( \dec{\bPhi}^\top\dec{\bPhi} \enc{\bm{m}}\enc{\bm{m}}^\top \right)
    \\
    \;&=\;
    \Tr\left( \dec{\bPhi}^\top\dec{\bPhi} \Cov_q[\bz] \right)
    \,+\,
    \enc{\bm{m}}^\top \dec{\bPhi}^\top\dec{\bPhi} \enc{\bm{m}}~.
\end{aligned}
\end{equation}

We substitute the quadratic term back into the full expression:
\begin{equation}
\begin{aligned}
    \cL_\text{recon}
    \;&=\;
    \bx^\top\bx
    \,-\, 2\bx^\top\dec{\bPhi}\enc{\bm{m}}
    \,+\, \enc{\bm{m}}^\top \dec{\bPhi}^\top\dec{\bPhi} \enc{\bm{m}}
    \,+\, \Tr\left( \dec{\bPhi}^\top\dec{\bPhi} \Cov_q[\bz] \right)
    \\
    \;&=\;
    (\bx - \dec{\bPhi}\enc{\bm{m}})^\top (\bx - \dec{\bPhi}\enc{\bm{m}})
    \,+\, \Tr\left( \dec{\bPhi}^\top\dec{\bPhi} \Cov_q[\bz] \right)
    \\
    \;&=\;
    \norm{\bx - \dec{\bPhi}\enc{\bm{m}}}_2^2
    \,+\, \Tr\left( \dec{\bPhi}^\top\dec{\bPhi} \Cov_q[\bz] \right)~.
\end{aligned}
\end{equation}

Therefore, for a general (non-factorized) posterior, the closed for linear decoder reconstruction loss is given by:
\begin{equation}
\boxed{
    \cL_\text{recon}
    \;=\;
    \norm{\bx - \dec{\bPhi}\enc{\bm{m}}}_2^2
    \,+\, \Tr\left( \dec{\bPhi}^\top\dec{\bPhi} \Cov_\enc{q}[\bz] \right)
}
\end{equation}

Assuming a factorized posterior (mean-field approximation), the covariance matrix $\Cov_q[\bz]$ is diagonal, with diagonal elements given by the variance vector $\enc{\bv} = \Var_q[\bz]$. The trace of the product of a matrix $\mathbf{A}$ and a diagonal matrix $\mathbf{D}$ is the dot product of their diagonals: $\Tr(\mathbf{A}\mathbf{D}) = \sum_i A_{ii} D_{ii}$.
Therefore:
\begin{equation}
    \Tr\left( \dec{\bPhi}^\top\dec{\bPhi} \Cov_q[\bz] \right)
    \;=\;
    \diag(\dec{\bPhi}^\top\dec{\bPhi})^\top \enc{\bv}~.
\end{equation}

This yields the final closed-form expression used in the text:
\begin{equation}
\boxed{
    \cL_\text{recon}
    \;=\;
    \underbracegray{\norm{\bx - \dec{\bPhi}\enc{\bm{m}}}_2^2}{\text{mean penalty}}
    \,+\,
    \underbracegray{\diag(\dec{\bPhi}^\top\!\dec{\bPhi})^\top\enc{\bv}}{\text{variance penalty}}
}
\end{equation}

\section{Derivation of the rectified Gaussian moments}
\label{sec:app:rectified_moments}
In this section, we derive the exact analytical expressions for the first and second moments of a Rectified Gaussian distribution. These moments are required to compute the closed-form reconstruction loss for the \greluvae (Table~\ref{tab:model_comparison}).

Let a single latent variable $z$ follow a Gaussian approximate posterior:
\begin{equation}
    q_\enc{\blambda}(z \vert \bx) \;=\; \cN(z; \enc{\mu}, \enc{\sigma}^2)~.
\end{equation}

The rectified representation is given by $h = \mathrm{ReLU}(z) = \max(0, z)$. To evaluate the reconstruction term $\expec{q}{\log p(\bx \vert h)}$, we require the expected value $\enc{m} = \expec{q}{h}$ and the variance $\enc{v} = \Var_{q}[h]$.

\subsubsection{Setup and change of variables.}
We seek to compute the expectation of functions of $h$ over the Gaussian density of $z$. Since $h=0$ for $z < 0$, the integration domain is effectively restricted to $z \in [0, \infty)$.
We perform a change of variables to the standard normal distribution using:
\begin{equation}
    u \;=\; \frac{z - \enc{\mu}}{\enc{\sigma}}
    \;\;\implies\;\;
    z \;=\; \enc{\mu} + \enc{\sigma} u~,
    \quad
    dz \;=\; \enc{\sigma} du~.
\end{equation}
The integration lower bound $z=0$ maps to the standardized limit:
\begin{equation}
    0 \;=\; \enc{\mu} + \enc{\sigma} u
    \;\;\implies\;\;
    u \;=\; -\frac{\enc{\mu}}{\enc{\sigma}} \;\coloneqq\; -\enc{\zeta}~.
\end{equation}
where we have defined the standardized mean $\enc{\zeta} \coloneqq \enc{\mu} / \enc{\sigma}$. The standard normal PDF is denoted by $\phi(u) = \frac{1}{\sqrt{2\pi}}e^{-u^2/2}$, and the CDF by $\Phi(u)$.

\subsubsection{First Moment: \texorpdfstring{$\expec{}{h}$}{E[h]}.}
The expected activation is given by:
\begin{equation}
    \enc{m}
    \;=\;
    \int_{-\infty}^{\infty} \max(0, z) \, \cN(z; \enc{\mu}, \enc{\sigma}^2) \, dz
    \;=\;
    \int_{0}^{\infty} z \, \cN(z; \enc{\mu}, \enc{\sigma}^2) \, dz~.
\end{equation}
Substituting the standardized variables:
\begin{equation}
    \enc{m}
    \;=\;
    \int_{-\enc{\zeta}}^{\infty} (\enc{\mu} + \enc{\sigma} u) \, \phi(u) \, du
    \;=\;
    \enc{\mu} \underbrace{\int_{-\enc{\zeta}}^{\infty} \phi(u) \, du}_{\text{Term A}}
    \;+\;
    \enc{\sigma} \underbrace{\int_{-\enc{\zeta}}^{\infty} u \, \phi(u) \, du}_{\text{Term B}}~.
\end{equation}

\textbf{Term A (probability mass).} Due to the symmetry of the Gaussian, the area under the curve from $-\enc{\zeta}$ to $\infty$ is equivalent to the area from $-\infty$ to $+\enc{\zeta}$:
\begin{equation}
    \int_{-\enc{\zeta}}^{\infty} \phi(u) \, du \;=\; 1 - \Phi(-\enc{\zeta}) \;=\; \Phi(\enc{\zeta})~.
\end{equation}

\textbf{Term B (tail expectation).} We use the identity $\frac{d}{du}(-\phi(u)) = u \phi(u)$:
\begin{equation}
    \int_{-\enc{\zeta}}^{\infty} u \phi(u) \, du
    \;=\;
    \Big[ -\phi(u) \Big]_{-\enc{\zeta}}^{\infty}
    \;=\;
    0 - \big(-\phi(-\enc{\zeta})\big)
    \;=\;
    \phi(\enc{\zeta})~,
\end{equation}
where we used $\phi(\infty)=0$ and the symmetry $\phi(-x) = \phi(x)$.

Combining Terms A and B yields the first moment:
\begin{equation} \label{eq:app:moment1}
\boxed{
    \expec{}{h}
    \;=\;
    \enc{m}
    \;=\;
    \enc{\mu} \Phi(\enc{\zeta})
    \,+\,
    \enc{\sigma} \phi(\enc{\zeta})
}
\end{equation}

\subsubsection{Second Moment: \texorpdfstring{$\expec{}{h^2}$}{E[h^2]}.}
The second raw moment is computed similarly:
\begin{equation}
    \expec{}{h^2}
    \;=\;
    \int_{-\enc{\zeta}}^{\infty} (\enc{\mu} + \enc{\sigma} u)^2 \, \phi(u) \, du
    \;=\;
    \int_{-\enc{\zeta}}^{\infty} (\enc{\mu}^2 + 2\enc{\mu}\enc{\sigma} u + \enc{\sigma}^2 u^2) \, \phi(u) \, du~.
\end{equation}

We split this into three integrals:
\begin{equation}
    \expec{}{h^2}
    \;=\;
    \enc{\mu}^2 \underbrace{\int_{-\enc{\zeta}}^{\infty} \phi(u) \, du}_{\Phi(\enc{\zeta})}
    \;+\;
    2\enc{\mu}\enc{\sigma} \underbrace{\int_{-\enc{\zeta}}^{\infty} u \phi(u) \, du}_{\phi(\enc{\zeta})}
    \;+\;
    \enc{\sigma}^2 \underbrace{\int_{-\enc{\zeta}}^{\infty} u^2 \phi(u) \, du}_{\text{Term C}}~.
\end{equation}

\textbf{Term C (Integration by Parts).} To solve $\int u^2 \phi(u) du$, we use integration by parts. Let $w = u$ and $dv = u \phi(u) du$. Then $dw = du$ and $v = -\phi(u)$.
\begin{align}
    \int_{-\enc{\zeta}}^{\infty} u^2 \phi(u) \, du
    \;&=\;
    \Big[ -u \phi(u) \Big]_{-\enc{\zeta}}^{\infty} - \int_{-\enc{\zeta}}^{\infty} \big(-\phi(u)\big) \, du \\
    \;&=\;
    \Big( 0 - \big[ -(-\enc{\zeta})\phi(-\enc{\zeta}) \big] \Big) + \Phi(\enc{\zeta}) \\
    \;&=\;
    \Big( 0 - \enc{\zeta}\phi(\enc{\zeta}) \Big) + \Phi(\enc{\zeta}) \\
    \;&=\;
    \Phi(\enc{\zeta}) - \enc{\zeta} \phi(\enc{\zeta})~.
\end{align}

Substituting Term C back into the main equation:
\begin{align}
    \expec{}{h^2}
    \;&=\;
    \enc{\mu}^2 \Phi(\enc{\zeta})
    \;+\;
    2\enc{\mu}\enc{\sigma} \phi(\enc{\zeta})
    \;+\;
    \enc{\sigma}^2 \Big( \Phi(\enc{\zeta}) - \enc{\zeta} \phi(\enc{\zeta}) \Big) \\
    \;&=\;
    (\enc{\mu}^2 + \enc{\sigma}^2) \Phi(\enc{\zeta})
    \;+\;
    2\enc{\mu}\enc{\sigma} \phi(\enc{\zeta})
    \;-\;
    \enc{\sigma}^2 \left(\frac{\enc{\mu}}{\enc{\sigma}}\right) \phi(\enc{\zeta}) \\
    \;&=\;
    (\enc{\mu}^2 + \enc{\sigma}^2) \Phi(\enc{\zeta})
    \;+\;
    \enc{\mu}\enc{\sigma} \phi(\enc{\zeta})~.
\end{align}

Thus, the second raw moment is:
\begin{equation} \label{eq:app:moment2}
\boxed{
    \expec{}{h^2}
    \;=\;
    (\enc{\mu}^2 + \enc{\sigma}^2) \Phi(\enc{\zeta})
    \,+\,
    \enc{\mu}\enc{\sigma} \phi(\enc{\zeta})
}
\end{equation}

\subsubsection{Variance.}
Finally, the variance is obtained via the standard definition $\enc{v} \,=\, \expec{}{h^2} - \enc{m}^2$.

\section{Extended methodological details and supplementary results}
Here we provide additional experimental details. We trained 128 models in total, comprising 64 \pvae and 64 \greluvae instances. Each family was evaluated over the full $8 \times 8$ grid of latent dimensionalities K and KL weighting coefficients $\beta$ (\cref{eq:free_energy_beta}):
\begin{equation*}
\begin{aligned}
K \;&\in\;
{64, 128, 192, 256, 384, 512, 1024, 2048},
\\
\beta \;&\in\;
{0.01, 0.1, 0.5, 1.0, 1.5, 2.0, 4.0, 8.0}.
\end{aligned}
\end{equation*}

\paragraph{Dataset.}
We used whitened van Hateren natural image patches \citep{van1998independent} following the preprocessing protocol from the \pvae paper~\citep{vafaii2024pvae}. The dataset consists of $16 \times 16$ pixel grayscale image patches extracted from the van Hateren natural image scenes dataset \citep{van1998independent}. The preprocessing pipeline includes:
\begin{enumerate}[leftmargin=15mm,rightmargin=15mm]
    \item Random extraction of $16 \times 16$ patches from natural images
    \item Whitening filter: $R(f) = f \cdot \exp((f/f_0)^n)$ with $f_0=0.5$ and $n=4$
    \item Local contrast normalization with kernel size 13 and $\sigma=0.5$
    \item Z-score normalization across spatial dimensions
\end{enumerate}

This preprocessing ensures that the input statistics approximate natural scene statistics while removing low-frequency correlations, encouraging the model to learn sparse, localized features. For additional details on the dataset preparation and statistics, we refer readers to the original \pvae paper~\citep{vafaii2024pvae}.

\paragraph{Model architectures.}
In this paper, we trained linear \pvae and \greluvae models. Both architectures share identical input/output dimensions but differ in their latent parameterization and decoding nonlinearity. Below we describe the specifications for a model with latent dimensionality $K=512$.

\paragraph{Encoder.}
\begin{itemize}[leftmargin=15mm,rightmargin=15mm]
    \item Input $\bx$: $16 \times 16 = 256$ dimensional flattened image patches.
    \item \textbf{Poisson:} Single linear layer $\enc{\mathbf{W}}_{\text{enc}} \in \mathbb{R}^{512 \times 256}$. Output is $512$ dimensional log-rate residuals $\enc{\bu}$.
    \item \textbf{Gaussian:} Single linear layer $\enc{\mathbf{W}}_{\text{enc}} \in \mathbb{R}^{1024 \times 256}$. Output splits into $512$ dimensional mean residuals $\enc{\delta\bmu}$ and $512$ dimensional log-scale residuals $\enc{\delta\bv}$.
    \item No bias terms, no nonlinearities.
\end{itemize}

\paragraph{Sampling.}
\begin{itemize}[leftmargin=15mm,rightmargin=15mm]
    \item \textbf{Poisson:} Rates are computed as $\enc{\blambda} = \dec{\blambda}_0 \odot \exp(\enc{\bu})$. Samples are drawn $\bz \sim \pois(\bz; \enc{\blambda})$.
    \item \textbf{Gaussian:} Means are computed as $\enc{\bmu} = \dec{\bmu}_0 + \enc{\delta \bmu}$, and scales are computed as $\enc{\bsigma} = \exp(\log\dec{\bsigma}_0 + \enc{\delta\bv})$. Samples are drawn $\bz \sim \cN(\bz; \enc{\bmu}, \enc{\bsigma}^2)$.
\end{itemize}

\paragraph{Decoder.}
\begin{itemize}[leftmargin=15mm,rightmargin=15mm]
    \item Input: Sampled latent variables $\bz$.
    \item Single linear layer: $\dec{\bPhi} \in \mathbb{R}^{256 \times 512}$ (dictionary).
    \item \textbf{Poisson:} Reconstruction $\hat{\bx} = \dec{\bPhi} \bz$.
    \item \textbf{Gaussian:} Reconstruction $\hat{\bx} = \dec{\bPhi} \mathrm{relu}(\bz)$.
    \item No bias terms, no nonlinearities.
\end{itemize}

\paragraph{Poisson training details.}
We trained the \pvae models using the exponential arrival time (EAT) Poisson relaxation, with a cubic indicator approximation \citep{ibrahim2026hitchhikers}. We used a relaxation temperature of $\tau = 0.05$

\paragraph{Learnable prior distributions.}
To ensure a fair comparison, we learned the prior parameters for both \pvae and \greluvae. We initialized the \pvae prior as a uniform distribution in log-space, which corresponds to a scale-invariant Jeffrey's prior. We initialized the \greluvae prior as the standard normal distribution.

\paragraph{Training configuration.}
\begin{itemize}[leftmargin=15mm,rightmargin=15mm]
    \item \textbf{Optimizer}: Adamax
    \item \textbf{Learning rate}: $\eta = 0.005$ 
    \item \textbf{Batch size}: 1000 samples per batch
    \item \textbf{Epochs}: 3000 training epochs + 5 warmup epochs
    \item \textbf{Weight decay}: 0.0 (no explicit L2 regularization)
    \item \textbf{Gradient clipping}: Maximum gradient norm of 500
    \item \textbf{Learning rate schedule}: Cosine annealing over all training epochs
\end{itemize}

\paragraph{Compute details.}
We used NVIDIA RTX 6000 Ada Generation GPUs without mixed-precision training to ensure numerical stability. Training a single model takes approximately 2-3 hours. The full experimental sweep required training 128 models in total, which was parallelized across multiple GPUs using a custom sweep runner that manages GPU memory allocation and job scheduling.

\subsection{Supplementary figures}

\begin{figure*}[t!]
    \centering
    \includegraphics[width=0.8\textwidth]{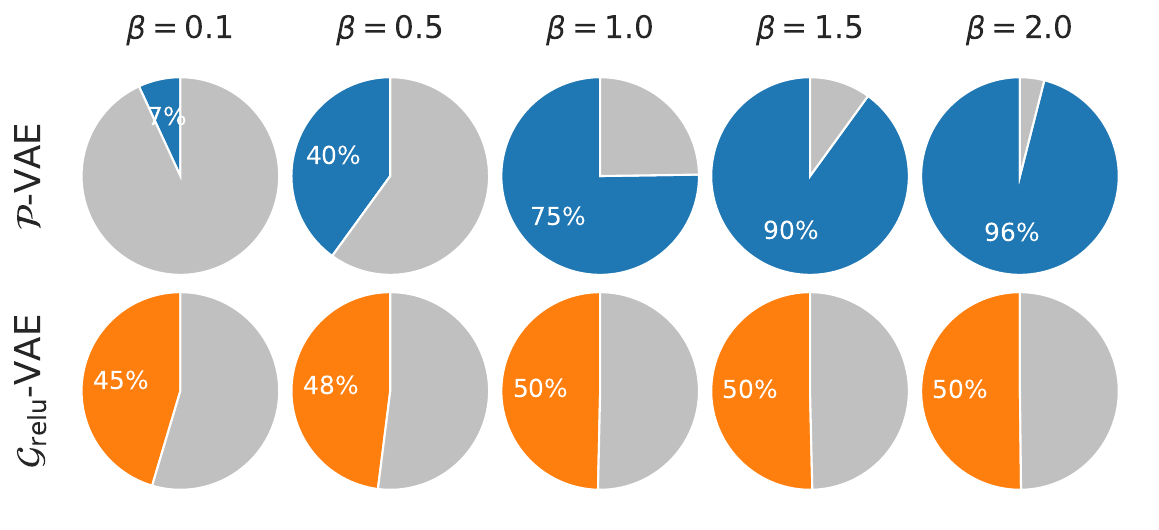}
    \caption{Proportion of zeros in learned representations as a function of $\beta$ (\cref{eq:free_energy_beta}). \textbf{Top:} \pvae exhibits tunable sparsity, with the fraction of zeros increasing from $7\%$ to $96\%$ as $\beta$ increases. \textbf{Bottom:} \greluvae remains near $50\%$ sparse irrespective of $\beta$, since the Gaussian prior does not penalize activation magnitude (compare \cref{eq:kl_poisson_approx,eq:kl_gaussian_approx}). Both models had $K=512$ latent dimensions. \Cref{fig:portion_zeros} replicates this result across a wide range of $K$.}
    \label{fig:pie}
\end{figure*}

\end{document}